\pdfoutput=1
\documentclass[11pt,onecolumn]{cleantechnicalreport}

\usepackage[authoryear,sort&compress,round]{natbib}
\usepackage{hybridfrontmatter}
\renewcommand{\ReportWebsiteIcon}{\faGithub}

\usepackage{bm}

\def\eqref#1{equation~\ref{#1}}

\def\1{\bm{1}}

\DeclareMathAlphabet{\mathsfit}{\encodingdefault}{\sfdefault}{m}{sl}
\SetMathAlphabet{\mathsfit}{bold}{\encodingdefault}{\sfdefault}{bx}{n}

\usepackage{makecell}
\usepackage{multirow}
\usepackage{adjustbox}
\usepackage{threeparttable}
\usepackage{float}
\usepackage[nameinlink,noabbrev]{cleveref}
\usepackage{subcaption}
\usepackage{wrapfig}
\usepackage{titletoc}

\definecolor{gray!0.9}{gray}{0.9}

\newtheorem{theorem}{Theorem}[section]

\newtheorem{definition}{Definition}[section]
\newtheorem{corollary}{Corollary}[section]

\usepackage[scr=boondoxo]{mathalpha}
\usepackage{xspace}
\newcommand{\model}{$\mathscr{c}$\texttt{mHC}\xspace}
\newcommand{\modelfull}{Contraction for Manifold-Constrained Hyper-Connections\xspace}

\renewcommand{\eqref}[1]{(\ref{#1})}

\title{\model: Balancing Stream Mixing in Hyper-Connections via Geometric Contraction}

\author[1,$^{*}$]{Haowei He}
\author[2,$^{*}$,$^{\dagger}$]{Ziwen Zhao}
\author[1,$^{*}$]{Chenshuo Pan}
\author[3,$^{\dagger}$]{Gang Lin}
\author[1]{Zhihao Yang}
\author[1]{Chao Wang}
\author[1]{Xinzhang Liu}
\author[1]{Kaidong Yu}
\author[4]{Han Gao}
\author[3]{Dongfang Li}
\author[2]{Menglin Yang}
\author[1,\faEnvelope]{\ReportAuthorBreak Shuangyong Song}
\author[1,\faEnvelope]{Yongxiang Li}
\affil[1]{\mbox{Xingchen AGI Lab, China Telecom}}
\affil[2]{\mbox{The Hong Kong University of Science and Technology (Guangzhou)}}
\affil[3]{\mbox{Harbin Institute of Technology (Shenzhen)}}
\affil[4]{\mbox{Huawei Technologies Co., Ltd.}}

\reportauthornote{$^{*}$ Equal contribution. hehw1@chinatelecom.cn, zzhao681@connect.hkust-gz.edu.cn, pancs@chinatelecom.cn \quad
$^{\dagger}$ Work done during an internship at Xingchen AGI Lab.\quad
\faEnvelope\ Corresponding author. songshy@chinatelecom.cn, liyx25@chinatelecom.cn
}
\reportwebsite{https://github.com/Newiz430/cmHC}

\begin{abstract}
Manifold-constrained Hyper-Connections (\texttt{mHC}) extends residual connections to multiple streams and stabilizes residual propagation by constraining stream-mixing residual matrices to the Birkhoff polytope, the set of doubly stochastic matrices.
In this work, we investigate \textit{how to effectively mix residual streams within the Birkhoff polytope}.
We reveal a \textbf{mixing trade-off}: when \textbf{under-mixed} residual matrices stay close to the vertices of the Birkhoff polytope, hyper-connections will degenerate into a set of nearly independent streams; when \textbf{over-mixed} residual matrices are concentrated around the uniform barycenter of the polytope, streams will become less distinguishable from each other. 
We theoretically validate our findings by bounding the cumulative residual mixing, stream difference and gradient difference of mixing coefficients across layers.
Motivated by these findings, we propose \modelfull (\model), an exact doubly stochastic method which geometrically contracts the Birkhoff polytope toward its interior.
This contraction mitigates under-mixing by forcing the residual matrices to explore deeper into the polytope, with a provable stream mixing bound and no new parameterization. 
Moreover, we apply a moderate contraction at the shallow layers to prevent over-mixing, enabling stable gradient propagation on extremely large-scale model pretraining. 
Extensive experiments across dense and mixture-of-experts models at various parameter scales demonstrate the effectiveness and scalability of \model.
We also develop fused operators for \model on Ascend NPUs, reducing hyper-connection execution latency by up to $44.8\%$.
\end{abstract}

\begin{document}

\maketitle

\section{Introduction}

\begin{figure*}[t]
  \centering
  \includegraphics[width=0.95\textwidth]{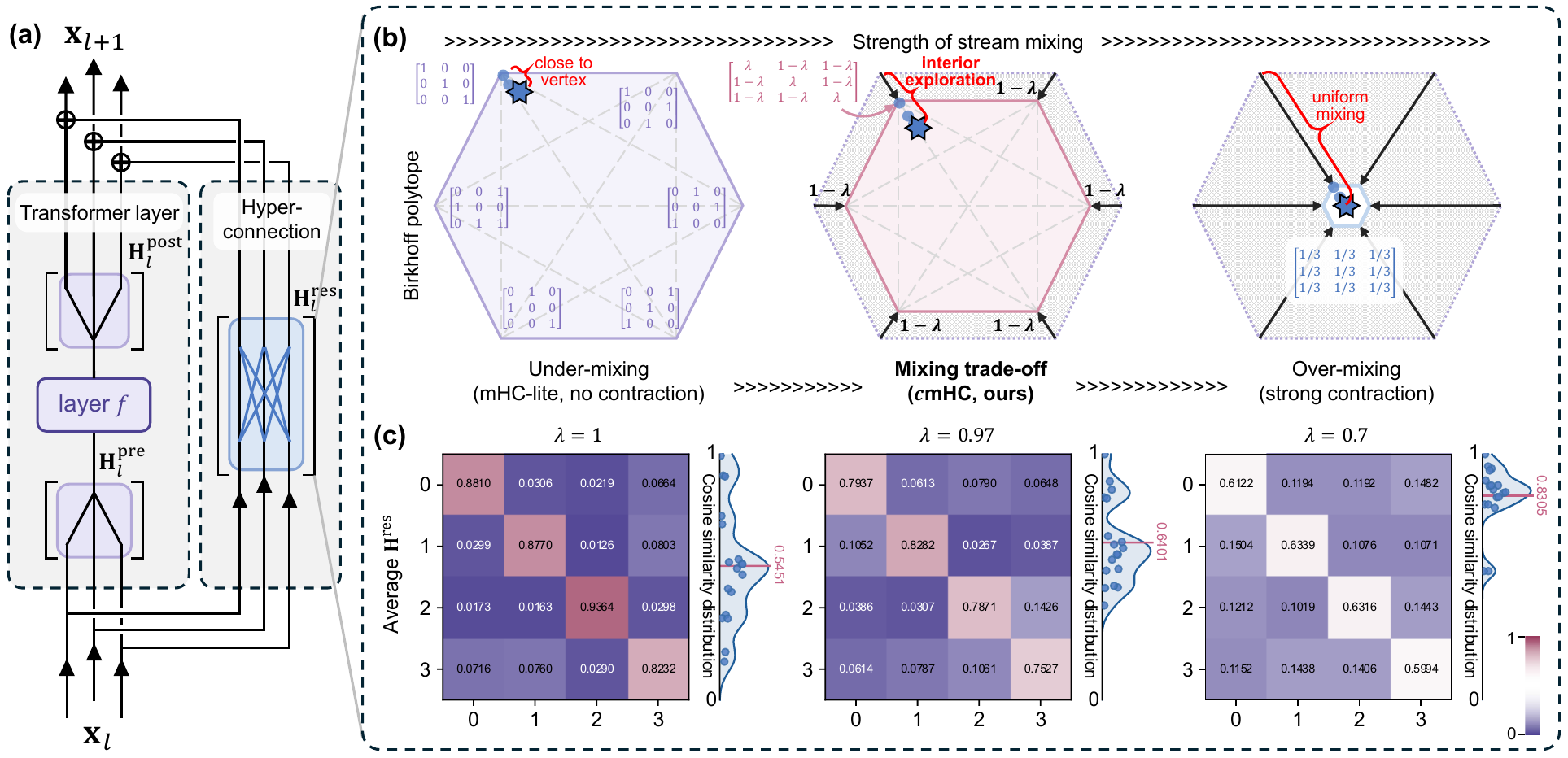}
  \caption{Illustration of our geometric contraction and its effect on stream mixing. 
\textbf{(a)} Architecture of hyper-connections at layer $l$. Our method focuses on the hyper-connection module.
\textbf{(b)} Two-dimensional projections of the Birkhoff polytope for $n\!=\!3$, along with the sampled trajectories of layer-averaged residual matrices during training, ended at the star~\smash{\raisebox{-2pt}{\includegraphics[scale=0.65]{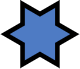}}}. Traditional methods like \texttt{mHC-lite} do not encourage an interior exploration of the polytope, so the residual matrices remain close to a permutation vertex; our moderate contraction encourages interior exploration, while a strong contraction may drive the residual matrices toward the uniform barycenter.
\textbf{(c)} Layer-averaged residual matrices and cosine similarity distributions between streams for $n\!=\!4$. A lower contraction coefficient $\lambda$ indicates a stronger contraction. %
}
\label{fig:birkhoff}
\end{figure*}

Hyper-Connections (\texttt{HC})~\citep{hc} extend standard residual connections~\citep{resnet} with learnable mixing across multiple feature streams, shown in Figure~\ref{fig:birkhoff}(a). 
As unconstrained mixing can destabilize signal propagation in deep networks, 
manifold-constrained Hyper-Connections (\texttt{mHC})~\citep{mhc} %
constrains the residual matrices to the Birkhoff polytope, the set of doubly stochastic matrices. \texttt{mHC-lite}~\citep{mhclite} provides an exact doubly stochastic constraint through learnable convex combinations of permutation matrices. 
Although these approaches exhibit training stability by doubly stochastic stream mixing, they do not ensure effective information exchange across streams. 
We study \textit{how to mix streams more effectively in hyper-connections} by analyzing two ill-conditioned mixing states: under-mixing and over-mixing. %
See Appendix~\ref{app:related-work} for more related work.

\paragraph{Under-mixing.}
Prior work has observed an under-mixing state in hyper-connections: residual mixing matrices often remain close to the identity, limiting cross-stream interaction~\citep{streamcollapse}. We further observe that this behavior is not specific to the identity: mixing can remain concentrated near other permutation vertices favored by initialization. 
In these cases, the streams are more permuted than actively mixed, limiting the expressivity of multi-stream architectures.  %
To address this issue, existing methods have explored the interior of the Birkhoff polytope through different parameterizations of residual matrices~\citep{shc,gomhc,ohc}, but they often rely on more involved constructions without a strictly guaranteed mixing effect, and their effectiveness is not validated on large-scale model pretraining. 
This motivates our \textbf{geometric contraction} strategy with \textit{a provable stream mixing bound} and \textit{no new parameterization}: 
we restrict the access to the permutation vertices and force the residual matrices to explore deeper into the polytope.
We propose \modelfull (\model), which contracts the polytope into a smaller region inside it by replacing permutation matrices with smoothed ones. This strictly guarantees a positive mixing while preserving exact double stochasticity. As illustrated in Figure~\ref{fig:birkhoff}(b), the contraction prevents residual mixing from degeneration around the permutation vertices, indicating guaranteed interior exploration. %

\paragraph{The mixing trade-off.}
Another important aspect of our work is the revelation of the potential \textbf{over-mixing} issue that, to our knowledge, remains largely unexplored in recent efforts. 
Although they encourage active stream mixing by different constraints and parameterizations, we argue that 
\textbf{stronger mixing is not always better, because excessive mixing makes feature streams less distinguishable}. As shown in Figure~\ref{fig:birkhoff}(c), the average similarity between streams becomes closer to 1 when the contraction is strengthened. %
We theoretically show that the cumulative mixing bound depends jointly on contraction strength and the number of contracted layers. We further reveal the ``\textit{butterfly effect}'' of stream mixing: even a numerically small contraction affects the stream difference and gradient difference of mixing coefficients, thereby amplifying over-mixing in deep networks.
These findings motivate a simple but effective strategy in \model: we apply moderate contraction only at shallow layers, encouraging early information exchange while allowing near-identity mappings in deeper layers. We validate the effectiveness and scalability of \model by extensive experiments on GPT-2
and mixture-of-experts at both 6B and 300B parameter scales, showing how the trade-off design achieves a lower training loss while stabilizing gradient propagation on extremely large-scale model pretraining. Furthermore, despite no new parameterization required, we improve the efficiency of \model in industrial scenarios by developing fused operators on Ascend NPUs, reducing hyper-connection execution latency by up to $44.8\%$.
In summary, our contributions are threefold:
\begin{itemize}[leftmargin=12pt,itemsep=1pt,topsep=2pt]
    \item We theoretically and empirically reveal the mixing trade-off in \texttt{mHC}: under-mixing makes hyper-connections degenerate into a set of single streams, while over-mixing makes streams less distinguishable from each other. %
    \item We propose \model, an exact doubly stochastic method which performs a moderate geometric contraction to the Birkhoff polytope without introducing additional parameters. %
    \item We evaluate \model through extensive experiments from GPT-2
to 6B-MoE and 300B-MoE to validate its effectiveness and scalability,  
    and develop efficient fused operators on Ascend NPUs to support efficient execution.
\end{itemize}

\section{From Under-mixing to Geometric Contraction}

\paragraph{Preliminary.} Hyper-connections maintain $n$ feature streams
$\mathbf{x}_l \in \mathbb{R}^{n \times d}$ and update them as
\begin{equation} \label{eq:hc}
\mathbf{x}_{l+1}
=
\mathbf{H}^{\mathrm{res}}_l \mathbf{x}_l
+
{\mathbf{H}^{\mathrm{post}}_l}^{\top}
f(\mathbf{H}^{\mathrm{pre}}_l \mathbf{x}_l;\mathbf{\Theta}_l),
\end{equation}
where $\mathbf{H}^{\mathrm{res}}_l \in \mathbb{R}^{n \times n}$
mixes residual streams and
$\mathbf{H}^{\mathrm{pre}}_l,\mathbf{H}^{\mathrm{post}}_l
\in \mathbb{R}^{1 \times n}$ aggregate and redistribute
the feature streams, respectively.
To ensure exact double stochasticity, \texttt{mHC-lite}
parameterizes the residual matrix as a convex combination
of the $n!$ permutation matrices $\mathbf{P}_m$:
\begin{equation} \label{eq:mhclite}
\mathbf{H}^{\mathrm{res}}_l
=
\sum_{m=1}^{n!}\alpha_{l,m}\mathbf{P}_m,
\qquad
\boldsymbol{\alpha}_l
=
\mathrm{softmax}
\left(
a_l^{\mathrm{res}}\mathbf{x}'_l\mathbf{W}^{\mathrm{res}}_l
+
\mathbf{b}^{\mathrm{res}}_l
\right),
\end{equation}
where $\mathbf{x}'_l$ is the normalized, flattened feature streams, and $a_l^{\mathrm{res}}$, $\mathbf{W}^{\mathrm{res}}_l$,
and $\mathbf{b}^{\mathrm{res}}_l$ are learnable parameters. 
This exactly constrains the residual matrices in a Birkhoff polytope $\mathscr{B}_n=\operatorname{conv} \{ \mathbf{P}_m\}_{m=1}^{n!}$ because of the Birkhoff--von Neumann
theorem~\citep{birkhoffvonneumann1}; see Appendix~\ref{app:mhclite} for details.

\subsection{Under-mixing in Hyper-Connections}

Under-mixing has been widely considered a key bottleneck that limits the expressivity of hyper-connections. Intuitively, it makes the multi-stream architecture degenerate into a set of nearly independent streams. Prior work has observed this phenomenon in \texttt{mHC-lite}: in most layers, $\mathbf{H}^{\mathrm{res}}_l$ stays close to the identity matrix with only small off-diagonal mass, instead of learning active stream mixing~\citep{shc,streamcollapse}. Geometrically,  $\mathbf{H}^{\mathrm{res}}_l$ remains near the identity vertex of the Birkhoff polytope during training, as shown in Figure \ref{fig:birkhoff}(b). 

\begin{wrapfigure}[20]{r}{0.56\textwidth}
\centering
%
\includegraphics[width=0.56\textwidth]{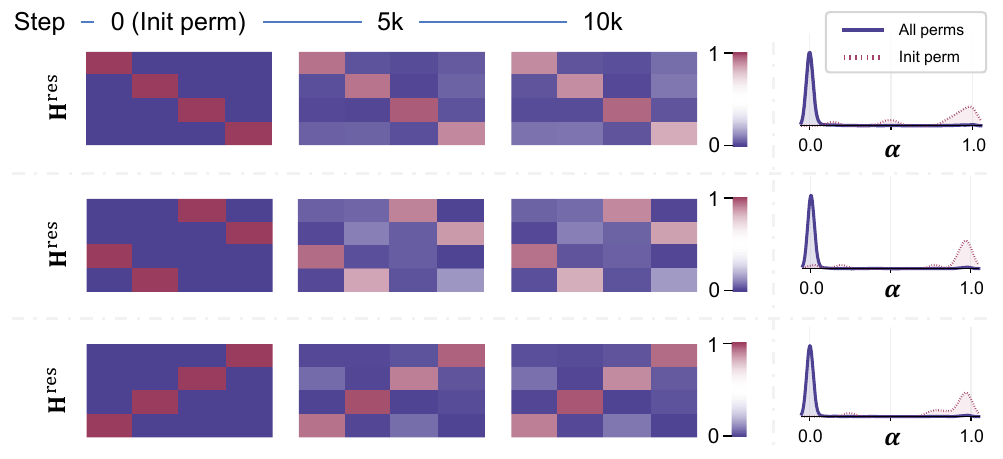}
\captionof{figure}{The under-mixing phenomenon under different initializations. \textbf{Left}: $\frac{1}{L}\sum_{l=0}^{L-1}\mathbf{H}^\mathrm{res}_l$  at training steps 0 (initial state), 5k, and 10k. 
Across different initializations, $\frac{1}{L}\sum_{l=0}^{L-1}\mathbf{H}^\mathrm{res}_l$ keeps close to the initial permutation matrix, not only the identity. 
\textbf{Right}: entry distributions of mixing coefficients $\boldsymbol{\alpha}$. 
The coefficient assigned to the initialized permutation (``Init perm'') stays close to 1, while the majority of coefficients over all permutations (``All perms'') stays close to 0.}
\label{fig:undermix}
\end{wrapfigure}

We further observe that \textbf{under-mixing is not specific to the identity; it can occur around any permutation matrix favored at initialization}. In \texttt{mHC-lite}, the input $\mathbf{x}_0$ is first duplicated across $n$ streams. Then, the diagonal entries of the residual bias $\mathbf{b}^{\mathrm{res}}_l$ are set to $0$, while the others are set to $-8$. This makes $\mathbf{H}^{\mathrm{res}}_l$ close to the identity matrix through the softmax in \eqref{eq:mhclite}.  When we change the initialization of $\mathbf{b}^{\mathrm{res}}_l$ to favor another permutation matrix, the average $\mathbf{H}^{\mathrm{res}}$ stays close to that permutation during training, as shown in Figure~\ref{fig:undermix}.

From the perspective of training dynamics, the stream $\mathbf{x}_{l,m}$ favored by initialization receives a larger coefficient update, which pushes its mixing coefficient $\alpha_{l,m}$ toward 1 and suppresses the remaining coefficients. As a result, a single $\mathbf{P}_m$ dominates $\mathbf{H}^{\mathrm{res}}_l$. Since $\mathbf{P}_m$ only permutes stream positions, cross-stream information exchange is suppressed. Figure~\ref{fig:undermix} shows that the entry distribution of $\boldsymbol{\alpha}$ remains close to its initial near-Dirac state during training: the coefficient of the initialized permutation stays near 1, while most other permutation coefficients stay near 0.%

\subsection{
    \texorpdfstring{%
        {\normalfont\model}%
    }{cmHC}:
    Geometric Contraction of the Birkhoff Polytope
}
\label{sec:contract}

To mitigate under-mixing, we introduce a geometric contraction $\mathscr{c}(\cdot;\lambda): \mathbb{R}^{n \times n} \rightarrow \mathbb{R}^{n \times n}$ which replaces each permutation matrix $\mathbf{P}_m$ with a smoothed version:
\begin{equation}
\label{eq:contract}
\mathbf{P}'_m = \mathscr{c}(\mathbf{P}_m;\lambda)
= \lambda\mathbf{P}_m + \frac{1-\lambda}{n-1}(\mathbf{1}_n\mathbf{1}_n^\top-\mathbf{P}_m).
\end{equation}
where $\lambda\in(1/n,1)$ controls the contraction strength. The case $\lambda \to 1$ recovers \texttt{mHC-lite}, while $\lambda\to1/n$ yields uniform mixing. The smoothed permutation matrices can be precomputed and replace the original bases without adding learnable parameters or changing their size.

Let $\mathbf{J}_n=\mathbf{1}_n\mathbf{1}_n^\top/n$ and $\lambda'=(n\lambda-1)/(n-1)$. Using the coefficients from \eqref{eq:mhclite}, the residual matrix is
\begin{equation}
\label{eq:contracthres}
\mathbf{H}^{\mathrm{res}'}_l
= \sum_{m=1}^{n!}\alpha_{l,m}\mathbf{P}'_m
= \mathscr{c}(\mathbf{H}^{\mathrm{res}}_l;\lambda)
= \lambda'\mathbf{H}^{\mathrm{res}}_l+(1-\lambda')\mathbf{J}_n.
\end{equation}
The resulting contraction polytope $\mathscr{B}_n(\lambda)=\operatorname{conv}\{\mathbf{P}'_m\}_{m=1}^{n!}$ has the following properties.

\begin{theorem}[Preservation of double stochasticity]
\label{thm:birkhoffproperty}
For $\lambda\in(1/n,1)$, $\mathscr{B}_n(\lambda)\subseteq\mathscr{B}_n$. In particular, every contracted residual matrix is doubly stochastic and has spectral norm at most one.
\end{theorem}

\begin{theorem}[Lower bound of stream mixing]
\label{thm:mixlbound}
For $\lambda\in(1/n,1)$, every entry of a contracted residual matrix satisfies
\begin{equation}
h^{\mathrm{res}'}_{l,ij}\ge \min \left( \lambda, \frac{1-\lambda}{n-1} \right) > 0.
\end{equation}
\end{theorem}

Proofs are provided in Appendices~\ref{app:birkhoffpropertyproof} and~\ref{app:mixlboundproof}.
Theorem~\ref{thm:birkhoffproperty} shows that \model preserves \textbf{exact double stochasticity}, ensuring $\|\mathbf{H}^{\mathrm{res}'}_l\|_2\le1$. Thus, for fixed mixing coefficients, the residual mixing operator and its transpose are non-expansive, preserving the stabilizing property of doubly stochastic mixing~\citep{mhc}. Meanwhile, Theorem~\ref{thm:mixlbound} guarantees a strictly positive contribution from every input stream to every output stream for $1/n<\lambda<1$, preventing extreme under-mixing. 

\paragraph{Contraction mapping.}
Equation~\eqref{eq:contracthres} expresses $\mathscr{c}$ as an interpolation between the original matrix and the uniform barycenter $\mathbf{J}_n$. For $1/n<\lambda<1$, we have $0<\lambda'<1$, and for any $\mathbf{H},\mathbf{G}\in\mathscr{B}_n$,
\begin{equation}
\label{eq:contraction-distance}
\|\mathscr{c}(\mathbf{H};\lambda)-\mathscr{c}(\mathbf{G};\lambda)\|_F
=\lambda'\|\mathbf{H}-\mathbf{G}\|_F.
\end{equation}
Thus, $\mathscr{c}$ is a contraction mapping on $\mathscr{B}_n$ under the Frobenius norm, with contraction factor $\lambda'$. Since $\mathscr{c}(\mathbf{J}_n;\lambda)=\mathbf{J}_n$, it also scales each matrix's distance to the barycenter by $\lambda'$, mapping the original polytope to a smaller region in its relative interior. The general definition of a contraction mapping is provided in Appendix~\ref{app:contraction-background}.

As illustrated in Figure~\ref{fig:birkhoff}(b), the original permutation vertices are no longer reachable. Even if $\boldsymbol{\alpha}_l$ concentrates on a single component, $\mathbf{H}^{\mathrm{res}'}_l$ becomes the smoothed $\mathbf{P}'_m$ rather than the permutation vertex $\mathbf{P}_m$. This indicates that the geometric contraction guarantees a positive cross-stream mixing, which mitigates the under-mixing effect.

\section{The Mixing Trade-off}

Stronger mixing promotes information exchange but may also reduce stream distinguishability. To examine this, we first analyze products of contracted residual matrices. We then study feature and gradient propagation using a local linear approximation to the Transformer branch, ${\mathbf{H}^\mathrm{post}}^\top f(\mathbf{H}^\mathrm{pre} \mathbf{x}; \mathbf{\Theta}) \rightarrow {\mathbf{H}^\mathrm{post}}^\top \mathbf{\Theta} \mathbf{H}^\mathrm{pre} \mathbf{x} = \mathbf{A}\mathbf{x}$, yielding
\begin{equation}
\label{eq:hclinear}
\mathbf{x}_{l+1}\approx(\mathbf{H}^{\mathrm{res}'}_l+\mathbf{A}_l)\mathbf{x}_l.
\end{equation}
Together, these analyses motivate a moderate contraction to shallow layers as a trade-off scheme.

\subsection{Convergence of Cumulative Contraction} \label{sec:cumulative}

Equation~\eqref{eq:contracthres} defines a contraction on the complete metric space $(\mathscr{B}_n(\lambda),\|\cdot\|_F)$ for $1/n<\lambda<1$. By Banach's fixed-point theorem~\citep{banach}, repeated application of this contraction converges to a unique fixed point, that is, the uniform barycenter $\mathbf{J}_n$. Convergence to the barycenter indicates an over-mixing state: \textbf{all output streams assign the same importance to every input stream, rather than learning dynamic stream interactions}. Details are provided in Appendix~\ref{app:contraction-background}.

Across network layers, the mixing effect accumulates through products of residual matrices. %
To quantify how contraction strength affects this cumulative mixing, we use \textit{Dobrushin's ergodicity coefficient}~\citep{dobrushin} $\tau(\mathbf{H}^{\mathrm{res}'}_l) \in [0,1]$ to measure the convergence of a non-homogeneous matrix product. It is defined as
\begin{equation}
\tau(\mathbf{H}^{\mathrm{res}'}_l) :=
1-\min_{i<j}\sum_{k=1}^{n}
\min(h^{\mathrm{res}'}_{l,ik},h^{\mathrm{res}'}_{l,jk}).
\end{equation}
A smaller $\tau$ means that different rows share more common mass; when $\tau$ approaches $0$, the rows become nearly identical, indicating stronger over-mixing. Then, the following theorem relates the convergence speed of cumulative stream mixing to the contraction strength.

\begin{theorem}[Convergence of cumulative stream mixing] \label{thm:dsproduct}
Let $\lambda_l\in(1/n,1]$ for $l=0,\ldots,L-1$, where $\lambda_l=1$ leaves layer $l$ uncontracted. Define $\lambda'_l=(n\lambda_l-1)/(n-1)$ and $\mathbf{H}^{\mathrm{res}'}_l=\lambda'_l\mathbf{H}^{\mathrm{res}}_l+(1-\lambda'_l)\mathbf{J}_n$. Then
\begin{equation}
\label{eq:cumulative-contraction-bound}
\tau\left(\prod_{l=0}^{L-1} \mathbf{H}^{\mathrm{res}'}_{L-1-l}\right)
\leq \prod_{l=0}^{L-1}\lambda'_l.
\end{equation}
\end{theorem}

\begin{wrapfigure}[19]{r}{0.55\textwidth}
\centering
\vspace{-12px}
\includegraphics[width=0.55\textwidth]{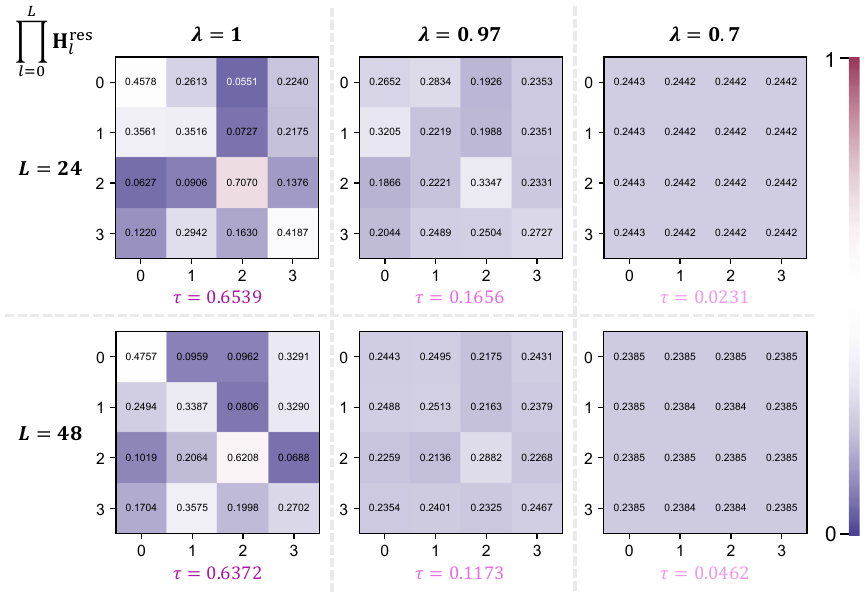}
\vspace{-20px}
\captionof{figure}{Cumulative residual matrix $\prod_{l=0}^{L-1} \mathbf{H}^{\mathrm{res}'}_l$ and Dobrushin's ergodicity coefficient $\tau$ under different $\lambda$ and network depth $L$. 
Note that the matrices slightly deviate from exact doubly stochasticity because of the inference-time low-precision conversion.}
\label{fig:prodmix}
\end{wrapfigure}

See Appendix~\ref{app:dsproductproof} for the proof. 
If all layers satisfy $\lambda'_l < 1$, the cumulative residual matrix approaches $\mathbf{J}_n$ \textbf{at an exponential rate} as $L$ increases. 
This implies that limiting such contraction to a small number of layers is important for alleviating the exponential convergence. 

Figure~\ref{fig:prodmix} visualizes the cumulative residual matrix $\prod_{l=0}^{L-1} \mathbf{H}^{\mathrm{res}'}_l$ and its Dobrushin coefficient $\tau$ under different $\lambda$ and $L$. When $\lambda$ decreases from $1$ to $0.97$, $\tau$ drops by about $0.5$, indicating a much stronger cumulative mixing effect. With $\lambda=0.7$, $\tau$ becomes close to $0$, and the cumulative residual matrix is nearly indistinguishable from the uniform barycenter of the Birkhoff polytope. And generally, increasing $L$ further reduces $\tau$, which is consistent with the decreasing cumulative upper bound. %

\subsection{The Butterfly Effect of Geometric Contraction} \label{sec:butterfly}

This section reveals a ``butterfly effect'' caused by geometric contraction: even a numerically small contraction can not only accelerate the convergence of the cumulative residual matrix, but also affect inference and gradient propagation of $\boldsymbol{\alpha}$, thereby increasing the risk of over-mixing in \model.

\paragraph{Stream difference $\|\tilde{\mathbf{x}}\|_F$ during inference.} To highlight the differences among streams, we first apply a mean-difference decomposition to the feature streams:
\begin{equation}
\mathbf{x} = \underbrace{\frac{1}{n}\mathbf{1}_n\mathbf{1}_n^\top \mathbf{x}}_{\text{stream mean}} + \underbrace{(\mathbf{I}_n-\frac{1}{n}\mathbf{1}_n\mathbf{1}_n^\top) \mathbf{x}}_{\text{stream difference } \tilde{\mathbf{x}}} = \mathbf{J}_n\mathbf{x} + \mathbf{\Pi}_n\mathbf{x},
\end{equation}
where $\mathbf{\Pi}_n=\mathbf{I}_n-\mathbf{J}_n$ and $\mathbf{I}_n$ is the $n \times n$ identity matrix. Every column of the stream difference $\tilde{\mathbf{x}}$ sums to $0$. 
To isolate how the contraction affects the stream differences, we align with the local linear approximation and do not consider the differences introduced by the non-linear Transformer branch: $\mathbf{\Pi}_n\mathbf{A}_l\mathbf{J}_n\approx\mathbf{0}$ for $l\ge1$\footnote{This only affects our additional over-mixing analysis below (Theorems~\ref{thm:streamdiffnorm} and~\ref{thm:alphagrad}), while the cumulative mixing bound in Theorem~\ref{thm:dsproduct} does not rely on this approximation.}.
We start from the output streams from the first layer $\tilde{\mathbf{x}}_1$ with stream difference norm $\Vert \tilde{\mathbf{x}}_1 \Vert_F >0$. 
The following theorem bounds the norm of the stream differences.

\begin{figure*}[t]
  \centering
  \includegraphics[width=\textwidth]{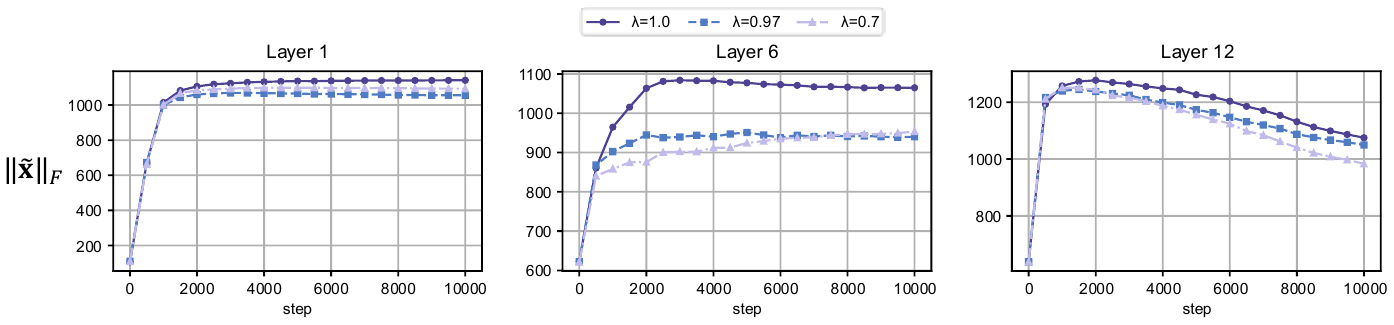}
  \caption{Frobenius norms of stream differences ${\Vert \tilde{\mathbf{x}}\Vert}_F$ over training steps. }
  \label{fig:streamnorm} \vspace{-0.5em}
\end{figure*}

\begin{theorem}[Norm upper bound of stream difference] \label{thm:streamdiffnorm}
For $L\ge2$ hyper-connection layers, the stream difference norm satisfies
\begin{equation}
\|\tilde{\mathbf{x}}_k\|_F \le \left(\prod_{l=1}^{k-1}(\lambda'_l+\ell_l)\right)\|\tilde{\mathbf{x}}_1\|_F
\end{equation}
for $2\le k\le L$, where $\lambda'_l = \frac{n\lambda_l-1}{n-1}$ is the normalized contraction coefficient and $\ell_l = \sup_{\mathbf{Z} \ne \mathbf{0}} \frac{\|\tilde{\mathbf{A}}_l \mathbf{Z}\|_F}{\| \mathbf{Z} \|_F}$ is the Lipschitz constant of $\tilde{\mathbf{A}}_l = \mathbf{\Pi}_n\mathbf{A}_l$.
\end{theorem}

See Appendix~\ref{app:streamdiffnormproof} for the detailed proof. 
Theorem~\ref{thm:streamdiffnorm} indicates that for fixed $\ell_l$ and $\tilde{\mathbf{x}}_1$, a stronger contraction can reduce the upper bound of the stream difference norm. 
The multiplicative accumulation implies that contraction in earlier layers can suppress differences at later layers. This reveals the over-mixing risk: stronger information exchange can make streams less distinguishable.

We analyze the norm of stream differences ${\Vert \tilde{\mathbf{x}} \Vert}_F$ in Figure~\ref{fig:streamnorm}. Note that the upper bound of ${\Vert \tilde{\mathbf{x}} \Vert}_F$ is also affected by the training stability of the Transformer layer ($\ell_l$), so deeper layers are not guaranteed to have smaller norms. Still, the stream difference norms for $\lambda=0.97$ and $\lambda=0.7$ are clearly lower than that of $\lambda=1$. This verifies that geometric contraction suppresses inter-stream differences.

\begin{wrapfigure}[16]{r}{0.4\textwidth}
\centering
\vspace{-0.25in}
\includegraphics[width=0.4\textwidth]{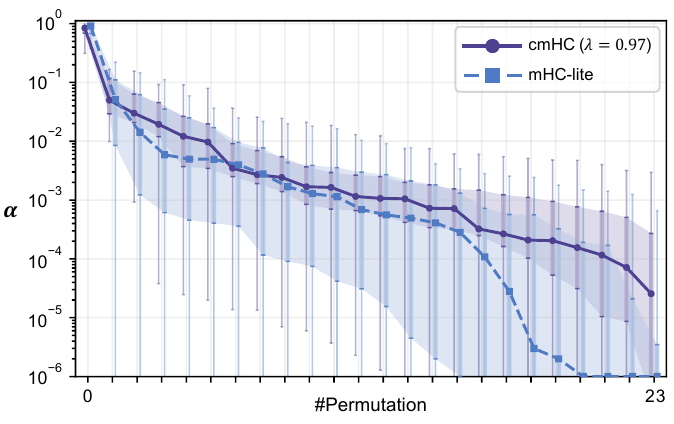}
\captionof{figure}{Distributions of mixing coefficients $\boldsymbol{\alpha}$ over the vertices of $\mathscr{B}_4$ (across all layers of nanoGPT-L). The $24$ vertices are ranked by their $\alpha$ value. Curves indicate the medians.
See Appendix~\ref{app:alpha} for full results.
}
\label{fig:alphadistrank}
\end{wrapfigure}

\paragraph{Gradient of the mixing coefficient $\boldsymbol{\alpha}$.} 
We next analyze how contraction strength affects the learning of mixing coefficients $\boldsymbol{\alpha}$. Intuitively, simply replacing the preset permutation matrices $\mathbf{P}_m$ with smoother versions $\mathbf{P}'_m$ should not directly affect the learning of $\boldsymbol{\alpha}$. However, we find that geometric contraction in \model induces a smoother $\boldsymbol{\alpha}$ distribution by reducing the upper bound of the gradient difference between any $\alpha$-pair of permutation matrices. Formally:

\begin{theorem}[Gradient difference of different permutations] \label{thm:alphagrad}
For two different permutation matrices $\mathbf P_i,\mathbf P_j \in \mathbb{R}^{n \times n}$, the gradient difference of the pretraining loss $\mathcal L$ w.r.t. their coefficients $\alpha_{l,i},\alpha_{l,j}$ is bounded by
\begin{equation}
\vert (\nabla_{\alpha_{l,i}}-\nabla_{\alpha_{l,j}})\mathcal L \vert \le 2\lambda'_l \Vert \nabla_{\tilde{\mathbf x}_{l+1}}\mathcal L \Vert_F \Vert \tilde{\mathbf x}_l \Vert_F.
\end{equation}
\end{theorem}

See Appendix~\ref{app:alphagradproof} for the detailed proof. 
From Theorem~\ref{thm:streamdiffnorm}, contraction lowers the bound on $\Vert \tilde{\mathbf x}_l \Vert_F$. For a fixed incoming gradient $\Vert \nabla_{\tilde{\mathbf x}_{l+1}}\mathcal L \Vert_F$, this also lowers the corresponding upper bound on $\vert (\nabla_{\alpha_{l,i}}-\nabla_{\alpha_{l,j}})\mathcal L \vert$. This reveals that contraction tends to assign similar gradients to different coefficients of permutation matrices, consequently \textbf{leading to a smoother distribution of $\boldsymbol{\alpha}$}, which further promotes stream mixing. 
This explains why a numerically small contraction can substantially encourage mixing, while a strong contraction may instead increase the risk of over-mixing.

To examine this effect, we sort the coefficients $\boldsymbol{\alpha}$ over all vertices of $\mathscr{B}_4$ in descending order, and compare their distributions across all layers between \texttt{mHC-lite} and \model with $\lambda=0.97$. 
Figure~\ref{fig:alphadistrank} shows that \model assigns more mass to the non-dominant vertices rather than only accumulating mass at vertex 0.
This supports that geometric contraction induces a smoother $\boldsymbol{\alpha}$ distribution.

\subsection{A Moderate Contraction to the Birkhoff Polytope}

The analyses in Secs.~\ref{sec:cumulative} and~\ref{sec:butterfly} identify a risk of over-mixing: cumulative contraction drives $\mathbf{H}^{\mathrm{res}'}$ exponentially toward the uniform barycenter of the Birkhoff polytope, and the contraction can also affect stream differences and induce a smoother coefficient distribution. 
These findings motivate a mixing trade-off strategy to control both strength and layers of the contraction. We \textbf{apply moderate contraction only to the first $K$ layers while allowing near-identity mappings in deeper layers}:
\begin{equation}
\mathbf{P}'_{l,m} = \begin{cases}
\lambda \mathbf{P}_m + \frac{1 - \lambda}{n-1} (\mathbf{1}_n\mathbf{1}_n^\top - \mathbf{P}_m), &l \le K-1, \\
\mathbf{P}_m, &l > K-1.
\end{cases}
\end{equation}
We set a single contraction coefficient $\lambda$ shared by all first $K$ layers. 
This improves the expressivity of hyper-connection architectures while reducing the risk of over-mixing. 

\paragraph{Why contract shallow layers?} 
Contractions at shallow layers are more effective than at deeper layers because \textbf{the contraction effect can accumulate and influence all subsequent layers.} We first provide a direct corollary from Theorem~\ref{thm:dsproduct} below to extend it to any prefix ending at layer $k$, where $1\leq k \leq L$.

\begin{corollary}[Cumulative mixing at intermediate layers]
\label{cor:dsproduct-prefix}
Let $\lambda_l\in(1/n,1]$ for $l=0,\ldots,L-1$. Then, for any $k\in\{1,\ldots,L\}$,
\begin{equation}
\label{eq:cumulative-contraction-bound-prefix}
\tau\left(\prod_{l=0}^{k-1} \mathbf{H}^{\mathrm{res}'}_{k-1-l}\right) \leq \prod_{l=0}^{k-1}\lambda'_l.
\end{equation}
\end{corollary}

This means the bound at layer $k$ depends only on the normalized contraction coefficients $\lambda_l'$ in the first $k$ layers. For a shallow contraction in the first $K$ layers, $\lambda_l<1$ for $l=0,\cdots,K$, so the Dobrushin's coefficient has a smaller bound at layer $k$. This shows that shallow contraction can propagate its mixing effect through subsequent intermediate layers\footnote{This longer-lasting effect could also increase the risk of over-mixing, so careful choices of $\lambda$ and $K$ are necessary.}, whereas deep contraction affects only the final stages of stream interaction. From a hierarchical perspective of deep representation learning, shallower layers also tend to encode more general features and deeper layers become more specialized~\citep{deep}, so it should be more cautious to mix streams in deeper layers. This finding is consistent with recent work highlighting the greater importance of early mixers~\citep{earlymixer}. We further provide empirical evidence in Sec.~\ref{sec:shallow}.

\section{Experiments}

\paragraph{Setup.}
Following~\cite{mhclite}, we evaluate language modeling on two extensive web datasets: OpenWebText~\citep{openwebtext} and FineWeb-Edu~\citep{finewebedu}. 
We compare \model with representative residual and hyper-connection baselines. (1) Vanilla residual connections ``Residual'' with one single stream $n=1$ and multiple independent streams (\textit{e.g., $n=4$}) without mixing. (2) Original \texttt{HC}~\citep{hc} without the Birkhoff constraint on the residual matrices. (3) \texttt{mHC}~\citep{mhc} with the Sinkhorn-Knopp iteration constraint. (4) \texttt{mHC-lite}~\citep{mhclite} with the exact Birkhoff von-Neumann parameterization of the residual matrices. (5) \texttt{KromHC}~\citep{kromhc} which decomposes doubly stochastic matrices into Kronecker products of smaller ones. See Appendix~\ref{app:setup} for more details.

\subsection{Results on GPT-2} \label{sec:expgpt}
\label{subsec:main_lm_results}

We train nanoGPT-S, nanoGPT-M, and nanoGPT-L from scratch for $10$k iterations using the nanoGPT framework~\citep{nanochat}. These models have 12, 24, and 48 hyper-connection layers\footnote{Each Transformer block contains two hyper-connection layers around attention and MLP layers, respectively.} and approximately $45$M, $0.12$B, and $0.36$B parameters, respectively. All methods use $n=4$ streams unless specified otherwise. Results are averaged over three runs.

\begin{table*}[t]
\centering
\caption{nanoGPT results on OpenWebText and FineWeb-Edu (mean $\pm$ standard deviation for 3 replicate experiments). All methods use 4 streams except ``Residual, $n=1$''. \textit{Avg. Loss Rank} marks the average rank over the Train Loss and Val. Loss. 
Best results are in \textbf{bold}. See Appendix~\ref{app:seed} for results on a different seed.}
\label{tab:main_lm_results}
\begin{threeparttable}
\setlength{\tabcolsep}{3.5pt}
\renewcommand{\arraystretch}{1.12}
\begin{adjustbox}{width=\textwidth,center}
\begin{tabular}{lcccccccccc}
\toprule[0.12em]
& \multicolumn{3}{c}{nanoGPT-S, \#Layers=12} & \multicolumn{3}{c}{nanoGPT-M, \#Layers=24} & \multicolumn{3}{c}{nanoGPT-L, \#Layers=48} &  \\
\cmidrule(lr){2-4}\cmidrule(lr){5-7}\cmidrule(lr){8-10}
Method & Train Loss & Val. Loss & Grad. Norm & Train Loss & Val. Loss & Grad. Norm & Train Loss & Val. Loss & Grad. Norm & \textit{Avg. Loss Rank} \\
\midrule
\rowcolor{gray!0.9} \multicolumn{11}{c}{\textit{OpenWebText}} \\
Residual, $n=1$ & $3.5677_{\pm 0.0006}$ & $3.5636_{\pm 0.0007}$ & $0.4239_{\pm 0.0194}$ & $3.3371_{\pm 0.0008}$ & $3.3363_{\pm 0.0007}$ & $0.3452_{\pm 0.0099}$ & $3.2311_{\pm 0.0009}$ & $3.2442_{\pm 0.0001}$ & $0.3330_{\pm 0.0064}$ & $11.8$ \\
Residual, $n=4$ & $3.4777_{\pm 0.0039}$ & $3.4742_{\pm 0.0036}$ & $0.3877_{\pm 0.0339}$ & $3.2591_{\pm 0.0029}$ & $3.2587_{\pm 0.0028}$ & $0.3751_{\pm 0.0150}$ & $3.1801_{\pm 0.0009}$ & $3.1909_{\pm 0.0011}$ & $0.4537_{\pm 0.0325}$ & $6.0$ \\ 
\texttt{HC} & $3.4749_{\pm 0.0017}$ & $3.4796_{\pm 0.0072}$ & $0.4300_{\pm 0.0584}$ & $3.2697_{\pm 0.0008}$ & $3.2681_{\pm 0.0006}$ & $0.4990_{\pm 0.1291}$ & $3.2232_{\pm 0.0013}$ & $3.2356_{\pm 0.0011}$ & $0.5519_{\pm 0.1373}$ & $9.7$ \\
\texttt{mHC} & $3.4813_{\pm 0.0059}$ & $3.4775_{\pm 0.0064}$ & $0.3882_{\pm 0.0269}$ & $3.2562_{\pm 0.0016}$ & $3.2558_{\pm 0.0013}$ & $0.3734_{\pm 0.0146}$ & $3.1795_{\pm 0.0020}$ & $3.1920_{\pm 0.0020}$ & $0.4665_{\pm 0.0357}$ & $6.0$ \\
\texttt{mHC-lite} & $3.4801_{\pm 0.0027}$ & $3.4759_{\pm 0.0021}$ & $0.3688_{\pm 0.0155}$ & $3.2578_{\pm 0.0034}$ & $3.2564_{\pm 0.0034}$ & $0.3854_{\pm 0.0197}$ & $3.1821_{\pm 0.0033}$ & $3.1949_{\pm 0.0033}$ & $0.4317_{\pm 0.0202}$ & $6.7$ \\
\texttt{KromHC} & $3.4809_{\pm 0.0024}$ & $3.4766_{\pm 0.0029}$ & $0.3898_{\pm 0.0338}$ & $3.2612_{\pm 0.0041}$ & $3.2602_{\pm 0.0041}$ & $0.3867_{\pm 0.0177}$ & $3.1807_{\pm 0.0019}$ & $3.1934_{\pm 0.0024}$ &  $0.4589_{\pm 0.0322}$ & $8.0$ \\
\rowcolor{violet!10}
\textbf{\model}, $\lambda=0.70$, all layers & $3.4790_{\pm 0.0045}$ & $3.4756_{\pm 0.0046}$ & $0.3637_{\pm 0.0134}$ & $3.2626_{\pm 0.0021}$ & $3.2618_{\pm 0.0023}$ & $0.3760_{\pm 0.0160}$ & $3.2063_{\pm 0.0016}$ & $3.2169_{\pm 0.0013}$ & $0.4346_{\pm 0.0164}$ & $9.0$ \\
\rowcolor{violet!10}
\textbf{\model}, $\lambda=0.85$, all layers & $3.4770_{\pm 0.0025}$ & $3.4732_{\pm 0.0026}$ & $0.3634_{\pm 0.0166}$ & $3.2579_{\pm 0.0031}$ & $3.2575_{\pm 0.0029}$ & $0.3758_{\pm 0.0150}$ & $3.2012_{\pm 0.0012}$ & $3.2119_{\pm 0.0014}$ & $0.4334_{\pm 0.0155}$ & $6.7$ \\
\rowcolor{violet!10}
\textbf{\model}, $\lambda=0.97$, all layers & $3.4745_{\pm 0.0014}$ & $3.4713_{\pm 0.0014}$ & $0.3629_{\pm 0.0121}$ & $3.2550_{\pm 0.0018}$ & $3.2542_{\pm 0.0020}$ & $0.3790_{\pm 0.0159}$ & $3.1915_{\pm 0.0016}$ & $3.2026_{\pm 0.0013}$ & $0.4456_{\pm 0.0252}$ & $4.2$ \\
\rowcolor{violet!10}
\textbf{\model}, $\lambda=0.97$, $K=10$ & $3.4755_{\pm 0.0035}$ & $3.4718_{\pm 0.0037}$ & $0.3805_{\pm 0.0195}$ & $3.2588_{\pm 0.0029}$ & $3.2584_{\pm 0.0029}$ & $0.3705_{\pm 0.0141}$ & $\mathbf{3.1785}_{\pm 0.0009}$ & $\mathbf{3.1893}_{\pm 0.0011}$ & $0.4340_{\pm 0.0150}$ & $4.2$ \\
\rowcolor{violet!10}
\textbf{\model}, $\lambda=0.97$, $K=5$ & $3.4731_{\pm 0.0028}$ & $3.4693_{\pm 0.0035}$ & $0.3726_{\pm 0.0182}$ & $3.2563_{\pm 0.0014}$ & $3.2549_{\pm 0.0017}$ & $0.3734_{\pm 0.0142}$ & $3.1801_{\pm 0.0007}$ & $3.1913_{\pm 0.0005}$ & $0.4341_{\pm 0.0141}$ & $3.2$ \\
\rowcolor{violet!10}
\textbf{\model}, $\lambda=0.97$, $K=2$ & $\mathbf{3.4723}_{\pm 0.0025}$ & $\mathbf{3.4686}_{\pm 0.0027}$ & $0.3774_{\pm 0.0200}$ & $\mathbf{3.2548}_{\pm 0.0006}$ & $\mathbf{3.2540}_{\pm 0.0007}$ & $0.3726_{\pm 0.0137}$ & $3.1791_{\pm 0.0032}$ & $3.1904_{\pm 0.0032}$ & $0.4353_{\pm 0.0140}$ & $\mathbf{1.3}$ \\

\midrule
\rowcolor{gray!0.9} \multicolumn{11}{c}{\textit{FineWeb-Edu}} \\
Residual, $n=1$ & $3.5284_{\pm 0.0023}$ & $3.5319_{\pm 0.0021}$ & $0.4395_{\pm 0.0204}$ & $3.3228_{\pm 0.0042}$ & $3.3206_{\pm 0.0042}$ & $0.3603_{\pm 0.0094}$ & $3.2308_{\pm 0.0002}$ & $3.2270_{\pm 0.0002}$ & $0.3347_{\pm 0.0059}$ & $11.2$ \\
Residual, $n=4$ & $3.4644_{\pm 0.0017}$ & $3.4684_{\pm 0.0017}$ & $0.3645_{\pm 0.0127}$ & $3.2489_{\pm 0.0005}$ & $3.2474_{\pm 0.0008}$ & $0.3689_{\pm 0.0136}$ & $3.1837_{\pm 0.0023}$ & $3.1819_{\pm 0.0023}$ & $0.4587_{\pm 0.0297}$ & $5.3$ \\ 
\texttt{HC} & $3.4699_{\pm 0.0001}$ & $3.4724_{\pm 0.0003}$ & $0.4493_{\pm 0.0708}$ & $3.2900_{\pm 0.0053}$ & $3.2898_{\pm 0.0056}$ & $2.7226_{\pm 2.2325}$ & $3.2413_{\pm 0.0188}$ & $3.2363_{\pm 0.0179}$ & $0.9607_{\pm 0.6755}$ & $10.8$ \\
\texttt{mHC} & $3.4626_{\pm 0.0009}$ & $3.4667_{\pm 0.0010}$ & $0.3684_{\pm 0.0125}$ & $3.2483_{\pm 0.0010}$ & $3.2471_{\pm 0.0005}$ & $0.3663_{\pm 0.0121}$ & $3.1862_{\pm 0.0009}$ & $3.1838_{\pm 0.0008}$ & $0.4755_{\pm 0.0338}$ & $4.0$ \\
\texttt{mHC-lite} & $\mathbf{3.4621}_{\pm 0.0008}$ & $\mathbf{3.4660}_{\pm 0.0013}$ & $0.3709_{\pm 0.0137}$ & $3.2480_{\pm 0.0026}$ & $3.2477_{\pm 0.0023}$ & $0.3742_{\pm 0.0163}$ & $3.1884_{\pm 0.0020}$ & $3.1846_{\pm 0.0021}$ & $0.4502_{\pm 0.0250}$ & $4.3$ \\
\texttt{KromHC} & $3.4643_{\pm 0.0016}$ & $3.4673_{\pm 0.0016}$ & $0.3653_{\pm 0.0123}$ & $3.2485_{\pm 0.0021}$ & $3.2480_{\pm 0.0018}$ & $0.3695_{\pm 0.0138}$ & $3.1836_{\pm 0.0006}$ & $\mathbf{3.1802}_{\pm 0.0004}$ & $0.4531_{\pm 0.0249}$ & $4.3$ \\
\rowcolor{violet!10}
\textbf{\model}, $\lambda=0.70$, all layers & $3.4678_{\pm 0.0020}$ & $3.4718_{\pm 0.0020}$ & $0.3648_{\pm 0.0118}$ & $3.2514_{\pm 0.0018}$ & $3.2503_{\pm 0.0021}$ & $0.3683_{\pm 0.0161}$ & $3.2067_{\pm 0.0010}$ & $3.2049_{\pm 0.0011}$ & $0.4320_{\pm 0.0014}$ & $9.3$ \\
\rowcolor{violet!10}
\textbf{\model}, $\lambda=0.85$, all layers & $3.4652_{\pm 0.0022}$ & $3.4691_{\pm 0.0019}$ & $0.3653_{\pm 0.0130}$ & $3.2466_{\pm 0.0017}$ & $3.2454_{\pm 0.0019}$ & $0.3750_{\pm 0.0174}$ & $3.2059_{\pm 0.0019}$ & $3.2043_{\pm 0.0019}$ & $0.4393_{\pm 0.0155}$ & $6.0$ \\
\rowcolor{violet!10}
\textbf{\model}, $\lambda=0.97$, all layers & $3.4673_{\pm 0.0022}$ & $3.4713_{\pm 0.0024}$ & $0.3593_{\pm 0.0118}$ & $3.2477_{\pm 0.0011}$ & $3.2463_{\pm 0.0009}$ & $0.3800_{\pm 0.0221}$ & $3.1936_{\pm 0.0013}$ & $3.1924_{\pm 0.0013}$ & $0.4644_{\pm 0.0346}$ & $6.3$ \\
\rowcolor{violet!10}
\textbf{\model}, $\lambda=0.97$, $K=10$ & $3.4679_{\pm 0.0004}$ & $3.4718_{\pm 0.0004}$ & $0.3682_{\pm 0.0137}$ & $\mathbf{3.2465}_{\pm 0.0019}$ & $\mathbf{3.2453}_{\pm 0.0018}$ & $0.3685_{\pm 0.0128}$ & $3.1883_{\pm 0.0023}$ & $3.1864_{\pm 0.0021}$ & $0.4563_{\pm 0.0290}$ & $5.7$ \\
\rowcolor{violet!10}
\textbf{\model}, $\lambda=0.97$, $K=5$ & $3.4642_{\pm 0.0008}$ & $3.4677_{\pm 0.0009}$ & $0.3651_{\pm 0.0125}$ & $3.2487_{\pm 0.0027}$ & $3.2474_{\pm 0.0024}$ & $0.3691_{\pm 0.0125}$ & $3.1845_{\pm 0.0024}$ & $3.1832_{\pm 0.0026}$ & $0.4556_{\pm 0.0253}$ & $5.0$ \\
\rowcolor{violet!10}
\textbf{\model}, $\lambda=0.97$, $K=2$ & $3.4634_{\pm 0.0002}$ & $3.4672_{\pm 0.0006}$ & $0.3683_{\pm 0.0136}$ & $3.2480_{\pm 0.0021}$ & $3.2464_{\pm 0.0020}$ & $0.3707_{\pm 0.0130}$ & $\mathbf{3.1834}_{\pm 0.0012}$ & $3.1817_{\pm 0.0014}$ & $0.4364_{\pm 0.0181}$ & $\mathbf{2.8}$ \\
\bottomrule[0.12em]
\end{tabular}
\end{adjustbox}
\end{threeparttable}
\end{table*}

\begin{figure*}[t]
  \centering
  \includegraphics[width=0.98\textwidth]{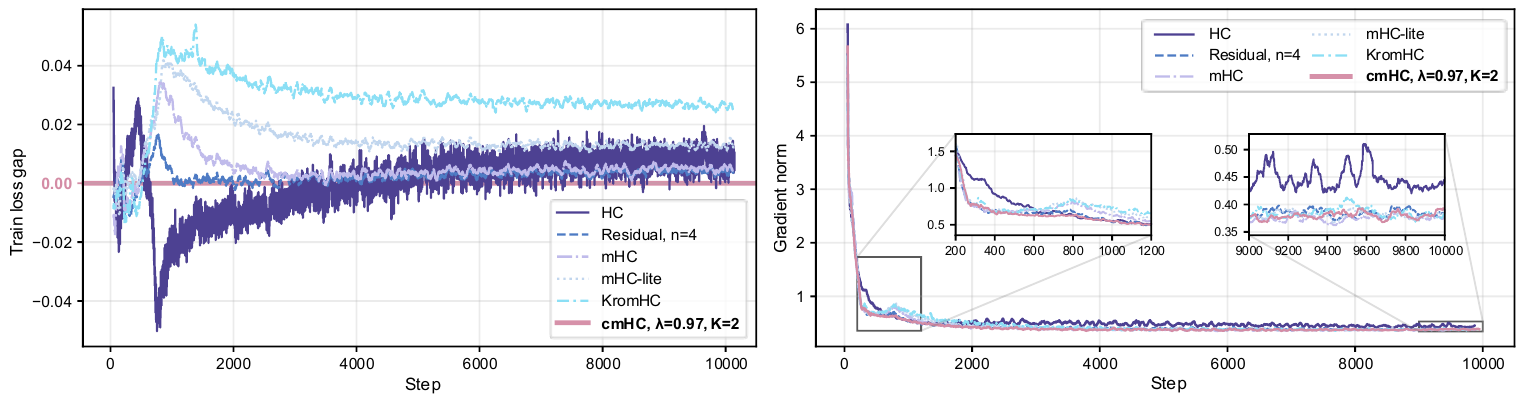}
  \caption{Loss gaps relative to \model and gradient norms (nanoGPT-S; OpenWebText). }
  \label{fig:lossgrad}
\end{figure*}

\paragraph{Main comparative results.}

Table~\ref{tab:main_lm_results} shows that \textbf{\model with $\lambda=0.97$ and $K=2$ achieves the best average loss rank on both datasets}. On OpenWebText, this configuration outperforms both \texttt{mHC} and \texttt{mHC-lite} in training and validation loss at every scale, reducing validation loss over \texttt{mHC-lite} by $0.0073$, $0.0024$, and $0.0045$, respectively. 
On FineWeb-Edu, \model achieves the lowest training and validation losses for nanoGPT-M ($K=10$) and the lowest training loss for nanoGPT-L ($K=2$), while remaining competitive on nanoGPT-S. Besides, the gradient norms of \model remain comparable to the traditional \texttt{mHC} and \texttt{mHC-lite} while significantly better than the unconstrained \texttt{HC}. These results demonstrate improved language modeling through moderate stream mixing without additional learnable parameters. 

\paragraph{Convergence and stability.} 

Figure~\ref{fig:lossgrad} shows the training curves of the loss gap and gradient norm. \model obtains the lowest training loss at convergence. Although \texttt{HC} appears to converge faster in the first 4k steps, its residual matrices are unconstrained, resulting in the largest training fluctuations among all methods. A similar pattern is observed in the gradient norm: \texttt{HC} continues to fluctuate by $\sim\!0.05$ in later training stages and remains substantially higher than the other baselines, whereas \model maintains one of the lowest and most stable gradient norms. Moreover, in Figure~\ref{fig:lossgrad}, all existing baselines have higher training losses than traditional residual connections with 4 independent streams, while \model is the only method that achieves a lower training loss.
See Appendix~\ref{app:gpt2} for more results on nanoGPT-M and nanoGPT-L.

\paragraph{Ablation on $\lambda$.}

With contraction applied to all layers, the moderate contraction $\lambda=0.97$ consistently outperforms the strong contraction $\lambda=0.70$ on both datasets. The validation-loss advantage grows from $0.0043$ to $0.0143$ on OpenWebText and from $0.0005$ to $0.0125$ on FineWeb-Edu as depth increases from 12 to 48 layers. These results support our mixing trade-off strategy that applies a moderate contraction.

\paragraph{Ablation on $K$.}

Fixing $\lambda=0.97$, we test the contraction on the first $K\in\{2,5,10\}$ and all layers. 
The results in Table~\ref{tab:main_lm_results} suggest that applying contraction to only a few layers works better than applying it to all layers, especially in deeper networks.
Average loss ranks also improve as $K$ decreases from $10$ to $5$ to $2$. These results support our mixing trade-off strategy that encourages stream mixing on shallow layers.

\begin{table*}[t]
\centering
\caption{Shallow and deep contractions on OpenWebText (mean ± standard deviation over 3 runs) with fixed $\lambda=0.97$. Best results are in \textbf{bold}.}
\label{tab:shallow_vs_deep}
\begin{threeparttable}
\setlength{\tabcolsep}{3.5pt}
\renewcommand{\arraystretch}{1.12}
\begin{adjustbox}{width=\textwidth,center}
\begin{tabular}{lcccccccccc}
\toprule[0.12em]
& \multicolumn{3}{c}{nanoGPT-S, \#Layers=12} & \multicolumn{3}{c}{nanoGPT-M, \#Layers=24} & \multicolumn{3}{c}{nanoGPT-L, \#Layers=48} & \\
\cmidrule(lr){2-4}\cmidrule(lr){5-7}\cmidrule(lr){8-10}
Method & Train Loss & Val. Loss & Grad. Norm & Train Loss & Val. Loss & Grad. Norm & Train Loss & Val. Loss & Grad. Norm & \textit{Avg. Loss Rank} \\
\midrule
\rowcolor{violet!10}
\textbf{\model}, $K=2$ & $\mathbf{3.4723}_{\pm 0.0025}$ & $\mathbf{3.4686}_{\pm 0.0027}$ & $0.3774_{\pm 0.0200}$ & $3.2548_{\pm 0.0006}$ & $3.2540_{\pm 0.0007}$ & $0.3726_{\pm 0.0137}$ & $\mathbf{3.1791}_{\pm 0.0032}$ & $\mathbf{3.1904}_{\pm 0.0032}$ & $0.4353_{\pm 0.0140}$ & $\mathbf{1.3}$ \\
\textbf{\model}, $K=-1$ & $3.4780_{\pm 0.0028}$ & $3.4743_{\pm 0.0027}$ & $0.3770_{\pm 0.0199}$ & $3.2557_{\pm 0.0019}$ & $3.2549_{\pm 0.0020}$ & $0.3728_{\pm 0.0148}$ & $3.1812_{\pm 0.0023}$ & $3.1918_{\pm 0.0026}$ & $0.4360_{\pm 0.0153}$ & $3.0$ \\
\textbf{\model}, $K=-2$ & $3.4847_{\pm 0.0022}$ & $3.4802_{\pm 0.0022}$ & $0.3940_{\pm 0.0337}$ & $3.2576_{\pm 0.0047}$ & $3.2569_{\pm 0.0046}$ & $0.3738_{\pm 0.0167}$ & $3.1795_{\pm 0.0018}$ & $3.1908_{\pm 0.0018}$ & $0.4368_{\pm 0.0151}$ & $3.2$ \\
\textbf{\model}, $K=-10$ & $3.5225_{\pm 0.0496}$ & $3.5184_{\pm 0.0492}$ & $1.3792_{\pm 0.1961}$ & $\mathbf{3.2528}_{\pm 0.0019}$ & $\mathbf{3.2518}_{\pm 0.0017}$ & $0.3761_{\pm 0.0169}$ & $3.1798_{\pm 0.0022}$ & $3.1906_{\pm 0.0021}$ & $0.4449_{\pm 0.0172}$ & $2.5$ \\
\bottomrule[0.12em]
\end{tabular}
\end{adjustbox}
\end{threeparttable}
\end{table*}

\subsection{Shallow vs. Deep Contraction} \label{sec:shallow}

To examine whether shallow contraction outperform deeper contraction, we conduct an experiment by keeping $\lambda=0.97$ and the other experimental settings fixed, and applying contraction to the last $m$ layers, denoted by $K=-m$ for $m\in\lbrace1,2,10\rbrace$. 

We can observe from the results in Table~\ref{tab:shallow_vs_deep} that \textbf{shallow contraction yields lower training and validation losses than deep contraction on both nanoGPT-S and nanoGPT-L}. The only exception is $K=-10$ on nanoGPT-M, but the same setting substantially raises nanoGPT-S validation loss to $3.5184$. \textbf{Deep contraction models also have higher gradient norms in 8/9 cases, suggesting less stable training.} These results indicate that deep contraction does not yield a consistent and robust improvement over shallow contraction. 

\subsection{Scalability: From 6B to 300B Parameters}

\label{sec:large-scale}

\paragraph{Setup.}
We evaluate scalability using 6B and 300B parameter mixture-of-experts~(MoE)~\citep{moe} language models trained on FineWeb-Edu. Both adopt a DeepSeek-style architecture~\citep{deepseek} with multi-latent attention and sparse MoE layers, using four residual streams. We set the \model contraction coefficient to $\lambda=0.97$ for 6B and $\lambda=0.96$ for 300B. Full architecture, training, and contraction settings are provided in Appendix~\ref{app:large-scale-setup}.

\begin{figure*}[t]
  \centering
  \includegraphics[width=0.98\textwidth]{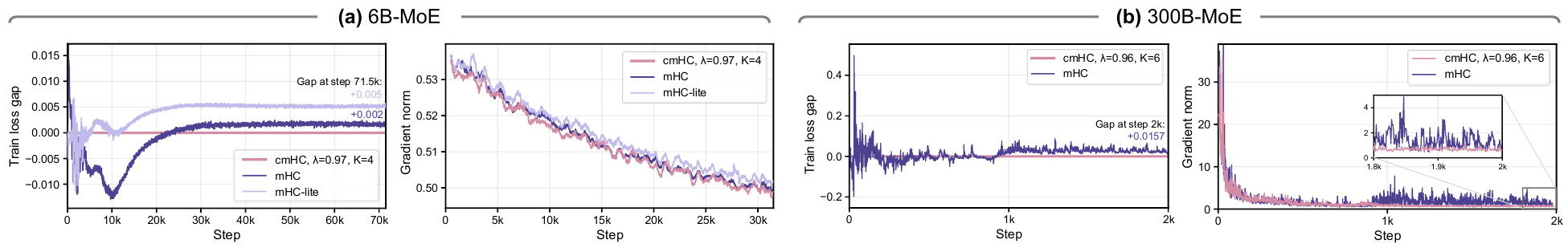}
  \caption{Loss gaps relative to \model and gradient norms (6B- and 300B-MoE; FineWeb-Edu).}
  \label{fig:large-scale-loss}
\end{figure*}

Figure~\ref{fig:large-scale-loss} compares the training loss gaps relative to \model and the corresponding gradient norms. At the 6B scale, after an early transient, \model finishes training with loss advantages of $0.002$ over \texttt{mHC} and $0.005$ over \texttt{mHC-lite} at step $71.5$k. 
Moreover, the gradient norm of \model keeps an advantage over the two baselines, indicating its training stability. 
From our scalability test on 300B-MoE, \model also attains a lower loss than \texttt{mHC}, with a gap of $0.0157$ at step $2$k; 
\model maintains smaller and smoother gradient norms, whereas \texttt{mHC} exhibits larger fluctuations. These results provide evidence that the benefit of our geometric contraction extends from 6B to substantially larger 300B models, including both lower losses and more stable optimization.

\subsection{Efficient Fused Operators on Ascend NPUs}
\label{sec:fused-operator-main}

Although the geometric contraction introduces no additional asymptotic computation,
a direct framework implementation still launches several small operators and repeatedly accesses the widened
multi-stream state. We therefore implement fused Pre and Post operators in Ascend C~\citep{huawei2025ascendc}. The Pre operator combines
coefficient generation, stream aggregation, and the projection of $\boldsymbol{\alpha}_l$ onto the contracted
permutation matrices $\{\mathbf{P}'_{l,m}\}_{m=1}^{n!}$; the Post operator fuses residual mixing and post expansion. Dense
projections run on cube cores, whereas reductions, nonlinearities, and the small-$n$ mixing operations run on
vector cores. The contracted bases are prepared once per layer and cached, so the kernels implement exactly the
same parameterization as \eqref{eq:contract}. Architecture diagrams, kernel data paths, backward equations, and
memory-layout details are provided in Appendix~\ref{app:fused-operator}.

We benchmark the complete fused-operator pipeline on an Ascend 910C server with CANN 9.1.0. All workloads use
$B=1$, $S=4096$, and $n=4$; the Transformer sublayer is excluded so that the comparison isolates the
hyper-connection operators. As shown in Table~\ref{tab:operator-main}, our implementation reduces total
forward $+$ backward latency by $44.8\%$, $31.0\%$, and $24.1\%$ as $D$ increases from $2,048$ to $7,168$. The benefit is
largest at smaller hidden dimensions, where launch and intermediate-memory overhead constitute a larger share
of execution time. Detailed Pre/Post breakdowns and comparison with an unfused MindSpore~\citep{mindspore2022whitepaper} implementation are
reported in Appendix~\ref{app:fused-operator}.

\begin{table}[t]
  \centering
  \small
  \caption{Total latency of the complete fused \model and \texttt{mHC} operator pipelines on Ascend 910C, including both forward and backward Pre and Post operators. Better latency time is in \textbf{bold}.}
  \label{tab:operator-main}
  \begin{tabular}{lcccc}
    \toprule[0.12em]
    \multirow{2}{*}{\makecell[c]{Input\\$[B,S,n,D]$}} & \multicolumn{2}{c}{Total latency ($\mu$s)} & \multirow{2}{*}{Speedup} & \multirow{2}{*}{Reduction} \\
     \cmidrule(){2-3}
      & \textbf{\model} & \texttt{mHC} & & \\
    \midrule
    $[1,4096,4,2048]$ & $\mathbf{1,474.59}$ & $2,670.02$ & $1.81\times$ & $44.8\%$ \\
    $[1,4096,4,3584]$ & $\mathbf{2,706.02}$ & $3,923.59$ & $1.45\times$ & $31.0\%$ \\
    $[1,4096,4,7168]$ & $\mathbf{6,472.56}$ & $8,523.01$ & $1.32\times$ & $24.1\%$ \\
    \bottomrule[0.12em]
  \end{tabular}
  \vspace{-0.3em}
\end{table}


%

%

%

\section{Conclusion}

We study how manifold-constrained hyper-connections mix information across feature streams. We first reveal that residual matrices in traditional methods tend to stay close to the permutation vertices of the Birkhoff polytope, leading to under-mixing, while excessive mixing can drive streams toward a uniform state and suppress their differences. Motivated by this mixing trade-off, we propose \model, which geometrically contracts the Birkhoff polytope toward its interior, providing a provable bound on stream mixing without introducing any new parameterization. Our analysis bounds cumulative residual mixing, stream differences and gradient differences of mixing coefficients across layers, showing that the effect of contraction depends jointly on its strength and the number of contracted layers. This motivates a simple but effective strategy that applies a moderate contraction to shallow layers to prevent over-mixing. Our comparative experiments and analyses of training dynamics from GPT-2 to 6B- and 300B-MoE models validate the effectiveness and scalability of \model. We further develop fused operators for \model on Ascend NPUs, reducing hyper-connection execution latency by up to 44.8\%, to further support industrial use.




\bibliographystyle{plainnat}
\bibliography{sample}

\newpage
\appendix
\startcontents[appendix]
\printcontents[appendix]{}{0}{\section*{Appendix Contents}} 

\newpage

\section{Preliminary} \label{app:preliminary}

\subsection{Manifold-Constrained Hyper-Connections (\texttt{mHC})} \label{app:mhc}

To resolve the training instability of \texttt{HC} caused by the unconstrained residual matrix $\mathbf{H}^{\mathrm{res}}$, DeepSeek proposed Manifold-Constrained Hyper-Connections (\texttt{mHC})~\citep{mhc} which constrains $\mathbf{H}^{\mathrm{res}}$ to the space of doubly stochastic matrices, known as the \textit{Birkhoff polytope} $\mathscr{B}_n$. A matrix $\mathbf{U} \in \mathbb{R}^{n \times n}$ is doubly stochastic if it satisfies:
\begin{equation}
    u_{ij} \geq 0 \ \forall i, j \in \{1, \dots, n\}, \quad \sum_{i=1}^n u_{ij} = 1 \ \forall i, \quad \sum_{j=1}^n u_{ij} = 1 \ \forall j.
\end{equation}
By forcing $\mathbf{H}^{\mathrm{res}}$ to be doubly stochastic, the mixing between feature streams becomes a convex combination. As the spectral norm of a doubly stochastic matrix is bounded by 1, it prevents signal amplification and guarantees the mathematical preservation of the identity mapping. In standard mHC, projection onto $\mathscr{B}_n$ is achieved via the iterative Sinkhorn-Knopp algorithm:
\begin{equation}
    \mathbf{H}^{\mathrm{res}}_{(0)} = \exp(\mathbf{S}), \quad \mathbf{H}^{\mathrm{res}}_{(t+1)} = \mathrm{Diag}\left(\mathbf{H}^{\mathrm{res}}_{(t)} \mathbf{1}_n\right)^{-1} \mathbf{H}^{\mathrm{res}}_{(t)} \mathrm{Diag}\left(\mathbf{1}_n^\top \mathbf{H}^{\mathrm{res}}_{(t)}\right)^{-1},
\end{equation}
where $\mathbf{S} \in \mathbb{R}^{n \times n}$ is a learned logit matrix. 

The components of a \texttt{mHC} block can be formalized as
\begin{equation} \label{eq:mhc}
\left\{
\begin{aligned}
\mathbf{x}'_l
&= \mathrm{RMSNorm}(\mathrm{flatten}(\mathbf{x}_l)), \\
\mathbf{H}^{\mathrm{pre}}_l
&= \sigma\left(a^{\mathrm{pre}}_l \mathbf{x}'_l \mathbf{W}^{\mathrm{pre}}_l + \mathbf{b}^{\mathrm{pre}}_l\right), \\
\mathbf{H}^{\mathrm{post}}_l
&= 2\sigma\left(a^{\mathrm{post}}_l \mathbf{x}'_l \mathbf{W}^{\mathrm{post}}_l + \mathbf{b}^{\mathrm{post}}_l\right), \\
\mathbf{H}^{\mathrm{res}}_l
&= \text{Sinkhorn-Knopp}\left(a^{\mathrm{res}}_l \mathrm{mat}\left(\mathbf{x}'_l \mathbf{W}^{\mathrm{res}}_l\right) + \mathbf{b}^{\mathrm{res}}_l\right),
\end{aligned}
\right.
\end{equation}
where $\mathbf{W}^{\mathrm{pre}}_l, \mathbf{W}^{\mathrm{post}}_l \in \mathbb{R}^{nd \times n}$, $\mathbf{W}^{\mathrm{res}}_l \in \mathbb{R}^{nd \times n^2}$ and $\mathbf{b}^{\mathrm{pre}}_l, \mathbf{b}^{\mathrm{post}}_l \in \mathbb{R}^{1\times n}$, $\mathbf{b}^{\mathrm{res}}_l \in \mathbb{R}^{1\times n^2}$ are linear projection weights and biases, respectively; $a^{\mathrm{pre}}_l, a^{\mathrm{post}}_l$ and $a^{\mathrm{res}}_l$ are learnable scalars; $\mathrm{mat}\!: \mathbb{R}^{1 \times n^2} \rightarrow \mathbb{R}^{n \times n}$ is a reshape map; $\sigma$ is the Sigmoid function.

\subsection{\texttt{mHC-lite}: Birkhoff-von Neumann Parameterization} \label{app:mhclite}

A known issue of \texttt{mHC}'s Sinkhorn-Knopp algorithm is that it does not guarantee exact double stochasticity. As it requires multiple iterations (typically $t \geq 10$ or $20$) to converge, the column sum of $\mathbf{H}^{\mathrm{res}}_l$ may deviate from 1 under insufficient iterations. 
To address this, \texttt{mHC-lite}~\citep{mhclite} learns a direct convex combination of a set of pre-defined permutation matrices $\mathbf{P}_m$:
\begin{equation}
    \mathbf{H}^\mathrm{res}_l = \sum_{m=1}^{n!} \alpha_{l,m} \mathbf{P}_m, \quad
    \boldsymbol{\alpha}_{l} = \mathrm{softmax}(a_l^\mathrm{res} \mathbf{x}'_l \mathbf{W}^{\mathrm{res}}_l + \mathbf{b}^{\mathrm{res}}_l),
\end{equation}
and other components are identical to \eqref{eq:mhc}. This strategy guarantees exact double stochasticity because of the Birkhoff-von Neumann Theorem:

\begin{theorem}[Birkhoff-von Neumann \citep{birkhoffvonneumann1,birkhoffvonneumann2}] \label{thm:birkhoff}
Any doubly stochastic matrix $\mathbf{U} \in \mathscr{B}_n$ can be represented as a convex combination of permutation matrices:
\begin{equation}
    \mathbf{U} = \sum_{m=1}^{n!} \alpha_m \mathbf{P}_m,
\end{equation}
where $\boldsymbol{\alpha} \in [0,1]^{n!}$ is a probabilistic simplex vector and $\{\mathbf{P}_m\}_{m=1}^{n!} \in \mathbb{R}^{n \times n}$ are permutation matrices.
\end{theorem}

\subsection{Contraction Mapping and the Uniform Fixed Point}
\label{app:contraction-background}

We provide the formal definition of a contraction mapping
and establish the fixed point and convergence rate of
the geometric map used in \model.

\begin{definition}[Contraction mapping]
Let $(\mathscr{M},d)$ be a metric space.
A map $\mathscr{c}:\mathscr{M}\to\mathscr{M}$ is a contraction
if there exists $0\le k<1$ such that
\begin{equation}
d(\mathscr{c}(u),\mathscr{c}(v))
\le k\,d(u,v),
\qquad \forall u,v\in\mathscr{M}.
\end{equation}
\end{definition}

\begin{theorem}[Banach fixed-point theorem~\citep{banach}]
\label{thm:banach}
A contraction on a non-empty complete metric space has a
unique fixed point. Starting from any point in the space,
repeated application of the map converges to this fixed point.
\end{theorem}

\paragraph{Applicability to the Birkhoff polytope.}
The Birkhoff polytope $\mathscr{B}_n$ is non-empty and closed
in the finite-dimensional normed space
$(\mathbb{R}^{n\times n},\|\cdot\|_F)$, and is therefore complete.
For a fixed $\lambda\in(1/n,1)$, define
\begin{equation}
\mathbf{J}_n=\frac{1}{n}\mathbf{1}_n\mathbf{1}_n^\top,
\qquad
\lambda'=\frac{n\lambda-1}{n-1}\in(0,1).
\end{equation}
Equation~\eqref{eq:contracthres} gives
$\mathscr{c}(\mathbf{H};\lambda)
=\lambda'\mathbf{H}+(1-\lambda')\mathbf{J}_n$.
Since $\mathbf{H},\mathbf{J}_n\in\mathscr{B}_n$ and
$\mathscr{B}_n$ is convex, this map sends
$\mathscr{B}_n$ into itself.
Moreover,
\begin{equation}
\|\mathscr{c}(\mathbf{H};\lambda)
-\mathscr{c}(\mathbf{G};\lambda)\|_F
=
\lambda'\|\mathbf{H}-\mathbf{G}\|_F.
\end{equation}
Thus, all conditions of Banach's fixed-point theorem hold.

\paragraph{Unique fixed point.}
The fixed point $\mathbf{H}_*$ satisfies
\begin{equation}
\mathbf{H}_*
=
\lambda'\mathbf{H}_*+(1-\lambda')\mathbf{J}_n.
\end{equation}
Since $1-\lambda'>0$, this implies
$\mathbf{H}_*=\mathbf{J}_n$.
Therefore, the unique fixed point is the uniform barycenter
of the Birkhoff polytope.

\paragraph{Convergence under repeated contraction.}
Let $\mathbf{H}^{(0)}\in\mathscr{B}_n$ and define
$\mathbf{H}^{(t+1)}
=\mathscr{c}(\mathbf{H}^{(t)};\lambda)$.
Subtracting $\mathbf{J}_n$ gives
\begin{equation}
\mathbf{H}^{(t+1)}-\mathbf{J}_n
=
\lambda'(\mathbf{H}^{(t)}-\mathbf{J}_n).
\end{equation}
By induction,
\begin{equation}
\mathbf{H}^{(t)}
=
\mathbf{J}_n+(\lambda')^t
(\mathbf{H}^{(0)}-\mathbf{J}_n),
\qquad
\|\mathbf{H}^{(t)}-\mathbf{J}_n\|_F
=
(\lambda')^t\|\mathbf{H}^{(0)}-\mathbf{J}_n\|_F.
\end{equation}
Hence, repeated contraction converges geometrically to
$\mathbf{J}_n$, with stronger contraction
(smaller $\lambda$, hence smaller $\lambda'$)
yielding faster convergence.
At this limit, every output stream receives the same
uniform mixture of input streams.

\section{Related Work}
\label{app:related-work}

\paragraph{Advances in large language models.}
Recent progress in large language models has been driven by advances
in pretraining, post-training, model architecture, and training systems.
Representative open-weight model families, including
Llama~\citep{llama3}, Qwen~\citep{qwen3}, and
TeleChat~\citep{he2024telechat,wang2024telechat}, have improved
multilingual language modeling and instruction following through
large-scale pretraining and alignment with human preferences.
Beyond general language capabilities,
DeepSeek-R1~\citep{deepseekr1} and the TeleChat2, TeleChat2.5, and
T1 series~\citep{wang2025technical} demonstrate the role of
post-training and reinforcement learning in strengthening
mathematical reasoning and code generation.
At the architectural and systems levels, sparse mixture-of-experts
models and optimized distributed training have enabled further
scaling, as illustrated by Qwen3~\citep{qwen3},
DeepSeek-V4~\citep{deepseekv4}, and
TeleChat3-MoE~\citep{liu2025training}.
More recent efforts increasingly coordinate architectural design
with execution efficiency: DeepSeek-V4 combines compressed attention
with manifold-constrained hyper-connections, while
Xing4~\citep{Song2026TeleChatTX} jointly develops model architectures,
training infrastructure, and reasoning and agentic post-training
on domestic chips and frameworks.
Together, these developments motivate stable and efficient
architectural components for large-scale models, providing the
broader context for our study of residual-stream mixing
in hyper-connections.

\paragraph{Multi-stream residual connections.}
Hyper-Connections~\citep{hc} extend residual connections~\citep{resnet} with learnable aggregation, mixing, and redistribution across multiple streams. To stabilize residual propagation, \texttt{mHC}~\citep{mhc} constrains residual mixing to doubly stochastic matrices through Sinkhorn normalization, while \texttt{mHC-lite}~\citep{mhclite} uses convex combinations of permutation matrices for an exact doubly stochastic constraint. 
There are also improvements focusing on the efficiency of computing and scalability on more streams, such as Kronecker decomposition~\citep{kromhc}, tensorization~\citep{temper} and sparse stream update~\citep{xhc}.
We discuss the mixing effectiveness of manifold-constrained hyper-connections, as existing research has revealed their under-mixing phenomenon, also referred to as ``stream collapse'' \citep{streamcollapse} or ``identity degradation'' \citep{shc}; this harms the expressivity of multi-stream design.

\paragraph{Exploration of the Birkhoff polytope.}
A line of recent work explores \textit{expansions} of the Birkhoff polytope to address under-mixing. For example, \cite{shc} constrain the residual matrices into a spectral sphere which expands the doubly stochastic constraint to the negative domain.
\cite{ohc} constrain residual matrices to $SO(n)$ rotation groups and parameterizes $SO(4)$ using pairs of unit quaternions. 
However, such expansions cannot strictly guarantee a positive mixing effect inside the Birkhoff polytope without provable stream mixing bounds. 
Another line of work focuses on a wider feasible region inside the Birkhoff polytope to encourage interior exploration.
For example, \cite{streamcollapse} recommend a non-symmetric initialization to unbind residual matrices from the polytope vertices; 
\cite{tbpmhc} covers the polytope with recursive transport construction; 
\cite{gomhc} and \cite{jpmhc} introduce the orthogonal constraint to expand the spectral reach and ensure the stability of the gradient spectrum of hyper-connections. 
However, these approaches often rely on more involved constructions and parameterizations, such as Cayley Transform and per-stream scaling vectors, while our approach introduces NO additional parameters and specialized constructions at all.

Moreover, the effectiveness of most existing approaches is not validated on extremely large-scale model pretraining. To our knowledge, the largest parameter scale used for research-based manifold-constrained hyper-connections is 28B by~\cite{xhc}, while we conduct scalability tests on a 300B mixture-of-experts model to validate the effectiveness and training stability of our method.

\paragraph{Stream similarity and over-mixing.}
A recent empirical analysis~\citep{ablaterescue} shows that high representational similarity between streams does not imply functional interchangeability, as the non-redundant part of streams can still play asymmetric roles to model predictions. 
Our work examines a different question: we study stream differences through the convergence of residual mixing and the resulting risk of over-mixing. Besides, our concern is that residual mixing can drive streams toward \textit{identical states}, rather than merely making their representations \textit{similar}. When streams become nearly identical to each other, the redundancy between streams can be amplified and the informative asymmetry can be weakened.

\section{Proofs}

\subsection{Proof of Theorem \ref{thm:birkhoffproperty}}  \label{app:birkhoffpropertyproof}
\noindent\textbf{Theorem \ref{thm:birkhoffproperty}} (Preservation of double stochasticity).
{\it 
For $\lambda\in(1/n,1)$, $\mathscr{B}_n(\lambda)\subseteq\mathscr{B}_n$. In particular, every contracted residual matrix is doubly stochastic and has spectral norm at most one.
}

\begin{proof}
By the linearity of the convex combination, we have
\begin{align}
\mathscr{c}(\mathbf{H}^{\mathrm{res}}_l;\lambda) &= \sum_{m=1}^{n!} \alpha_{l,m} \left( \lambda \mathbf{P}_m + \frac{1 - \lambda}{n-1} (\mathbf{1}_n\mathbf{1}_n^\top - \mathbf{P}_m) \right) \\
&= \lambda \left( \sum_{m=1}^{n!} \alpha_{l,m} \mathbf{P}_m \right) + \frac{1 - \lambda}{n-1} \left( \mathbf{1}_n\mathbf{1}_n^\top - \sum_{m=1}^{n!} \alpha_{l,m} \mathbf{P}_m \right) \\
&= \lambda \mathbf{H}^{\mathrm{res}}_l + \frac{1 - \lambda}{n-1} (\mathbf{1}_n\mathbf{1}_n^\top - \mathbf{H}^{\mathrm{res}}_l)
\end{align}
for any doubly stochastic matrix $\mathbf{H}^{\mathrm{res}}_l$. 
As $\mathbf{H}^{\mathrm{res}}_l \mathbf{1}_n=\mathbf{1}_n$, $\mathbf{1}_n^\top \mathbf{H}^{\mathrm{res}}_l= \mathbf{1}_n^\top$, we have
\begin{align}
\mathscr{c}(\mathbf{H}^{\mathrm{res}}_l;\lambda) \mathbf{1}_n &= \lambda \mathbf{H}^{\mathrm{res}}_l  \mathbf{1}_n + \frac{1 - \lambda}{n-1} (\mathbf{1}_n\mathbf{1}_n^\top - \mathbf{H}^{\mathrm{res}}_l)  \mathbf{1}_n
=\left(\lambda - \frac{1 - \lambda}{n-1} + \frac{n(1 - \lambda)}{n-1} \right) \mathbf{1}_n = \mathbf{1}_n, \\
\mathbf{1}_n^\top\mathscr{c}(\mathbf{H}^{\mathrm{res}}_l;\lambda) &= \mathbf{1}_n^\top\lambda \mathbf{H}^{\mathrm{res}}_l + \frac{1 - \lambda}{n-1} \mathbf{1}_n^\top(\mathbf{1}_n\mathbf{1}_n^\top - \mathbf{H}^{\mathrm{res}}_l)
=\mathbf{1}_n^\top\left(\lambda - \frac{1 - \lambda}{n-1} + \frac{n(1 - \lambda)}{n-1} \right) = \mathbf{1}_n^\top.
\end{align}
So $\mathscr{c}(\mathbf{H}^{\mathrm{res}}_l;\lambda)$ is doubly stochastic and thus $\mathscr{B}_n(\lambda) \subseteq \mathscr{B}_n$. For any $\mathbf{H}^{\mathrm{res}'}_l \in \mathscr{B}_n(\lambda)$, $\Vert \mathbf{H}^{\mathrm{res}'}_l \Vert_1=\Vert \mathbf{H}^{\mathrm{res}'}_l \Vert_\infty=1$, so the spectral norm satisfies $\Vert \mathbf{H}^{\mathrm{res}'}_l \Vert \le \sqrt{\Vert \mathbf{H}^{\mathrm{res}'}_l \Vert_1\Vert \mathbf{H}^{\mathrm{res}'}_l \Vert_\infty}=1$.

\end{proof}

\subsection{Proof of Theorem \ref{thm:mixlbound}}  \label{app:mixlboundproof}

\noindent\textbf{Theorem \ref{thm:mixlbound}} (Lower bound of stream mixing). 
{\it 
For $\lambda\in(1/n,1)$, every entry of a contracted residual matrix satisfies
\begin{equation}
h^{\mathrm{res}'}_{l,ij}\ge \min \left( \lambda, \frac{1-\lambda}{n-1} \right) > 0.
\end{equation}
}

\begin{proof}
For each entry of $\mathbf{H}^{\mathrm{res}'}_l$, we have
\begin{align}
h^{\mathrm{res}'}_{l,ij} &= \sum_{m=1}^{n!} {\alpha_{l,m} \left(\lambda p_{m,ij} + \frac{1 - \lambda}{n-1} (1 - p_{m,ij})\right)} \\
&=\lambda\sum_{p_{m,ij}=1} {\alpha_{l,m}} + \frac{1-\lambda}{n-1} \sum_{p_{m,ij}=0} {\alpha_{l,m}}  \\
&=\lambda\sum_{p_{m,ij}=1} {\alpha_{l,m}} + \frac{1-\lambda}{n-1} \left(1-\sum_{p_{m,ij}=1} {\alpha_{l,m}} \right).
\end{align}
As $0 \le \sum_{p_{m,ij}=1} {\alpha_{l,m}} \le 1$, $h^{\mathrm{res}'}_{l,ij}$ is a convex combination of $\lambda$ and $\frac{1-\lambda}{n-1}$, hence $h^{\mathrm{res}'}_{l,ij} \ge \min \left( \lambda, \frac{1-\lambda}{n-1} \right)$.
\end{proof}

\subsection{Proof of Theorem \ref{thm:dsproduct}}  \label{app:dsproductproof}

\noindent\textbf{Theorem \ref{thm:dsproduct}} (Convergence of cumulative stream mixing). 
{\it 
Let $\lambda_l\in(1/n,1]$ for $l=0,\ldots,L-1$, where $\lambda_l=1$ leaves layer $l$ uncontracted. Define $\lambda'_l=(n\lambda_l-1)/(n-1)$ and $\mathbf{H}^{\mathrm{res}'}_l=\lambda'_l\mathbf{H}^{\mathrm{res}}_l+(1-\lambda'_l)\mathbf{J}_n$. Then
\begin{equation}
\label{eq:cumulative-contraction-bound-appendix}
\tau\left(\prod_{l=0}^{L-1} \mathbf{H}^{\mathrm{res}'}_{L-1-l}\right)
\leq \prod_{l=0}^{L-1}\lambda'_l.
\end{equation}
}

\begin{proof}

For $1/n<\lambda_l\le1$, the affine form of $\mathbf{H}^{\mathrm{res}'}_l$ and the nonnegativity of $\mathbf{H}^{\mathrm{res}}_l$ give
\begin{equation} \label{eq:hresentrybound}
h^{\mathrm{res}'}_{l,ij}
=\lambda'_l h^{\mathrm{res}}_{l,ij}+\frac{1-\lambda'_l}{n}
\geq \frac{1-\lambda_l}{n-1}\geq0.
\end{equation}
This includes $\lambda_l=1$, where the entrywise lower bound is zero and $\mathbf{H}^{\mathrm{res}'}_l=\mathbf{H}^{\mathrm{res}}_l$ remains doubly stochastic.

For every layer $l$, the entry-wise lower bound \eqref{eq:hresentrybound} gives
\begin{equation}
\sum_{k=1}^{n}
\min(h^{\mathrm{res}'}_{l,ik},h^{\mathrm{res}'}_{l,jk}) \geq \frac{n(1-\lambda_l)}{n-1}
\quad \Rightarrow \quad 
\tau(\mathbf{H}^{\mathrm{res}'}_l) \leq 1-\frac{n(1-\lambda_l)}{n-1}=\lambda'_l\leq1.
\end{equation}
Dobrushin's coefficient is sub-multiplicative ($\tau(\mathbf{A}\mathbf{B}) \leq \tau(\mathbf{A})\tau(\mathbf{B})$), so
\begin{equation}
\tau\left(\prod_{l=0}^{L-1} \mathbf{H}^{\mathrm{res}'}_{L-1-l}\right) \leq \prod_{l=0}^{L-1}\tau(\mathbf{H}^{\mathrm{res}'}_l) \leq \prod_{l=0}^{L-1}\lambda'_l.
\end{equation}

\end{proof}

\subsection{Proof of Theorem \ref{thm:streamdiffnorm}}  \label{app:streamdiffnormproof}

\noindent\textbf{Theorem \ref{thm:streamdiffnorm}} (Norm upper bound of stream difference). 
{\it 
For $L\ge2$ hyper-connection layers, the stream difference norm satisfies
\begin{equation}
\|\tilde{\mathbf{x}}_k\|_F \le \left(\prod_{l=1}^{k-1}(\lambda'_l+\ell_l)\right)\|\tilde{\mathbf{x}}_1\|_F
\end{equation}
for $2\le k\le L$, where $\lambda'_l = \frac{n\lambda_l-1}{n-1}$ is the normalized contraction coefficient and $\ell_l = \sup_{\mathbf{Z} \ne \mathbf{0}} \frac{\|\tilde{\mathbf{A}}_l \mathbf{Z}\|_F}{\| \mathbf{Z} \|_F}$ is the Lipschitz constant of $\tilde{\mathbf{A}}_l = \mathbf{\Pi}_n\mathbf{A}_l$.
}

\begin{proof}

Since $\mathbf{H}^{\mathrm{res}'}_l$ is doubly stochastic, $\mathbf{\Pi}_n\mathbf{H}^{\mathrm{res}'}_l=\mathbf{H}^{\mathrm{res}'}_l-\mathbf{J}_n=\mathbf{H}^{\mathrm{res}'}_l\mathbf{\Pi}_n$. Applying $\mathbf{\Pi}_n$ to \eqref{eq:hclinear} and substituting \eqref{eq:contracthres} gives
\begin{align} \label{eq:streamdiffiter}
\tilde{\mathbf{x}}_{l+1}
&\approx \mathbf{H}^{\mathrm{res}'}_l\tilde{\mathbf{x}}_l+\mathbf{\Pi}_n\mathbf{A}_l\mathbf{x}_l \\
&= \lambda'_l\mathbf{H}^{\mathrm{res}}_l\tilde{\mathbf{x}}_l
+\tilde{\mathbf{A}}_l\tilde{\mathbf{x}}_l
+\mathbf{\Pi}_n\mathbf{A}_l\mathbf{J}_n\mathbf{x}_l \\
&\approx (\lambda'_l\mathbf{H}^{\mathrm{res}}_l+\tilde{\mathbf{A}}_l)\tilde{\mathbf{x}}_l,
\end{align}
where $\tilde{\mathbf{A}}_l=\mathbf{\Pi}_n\mathbf{A}_l$ and $\lambda'_l = \frac{n\lambda_l-1}{n-1} \in (0,1)$. For \model with $L \ge 2$ hyper-connection layers, the stream difference for any output layer index $2 \le k \le L$ is
\begin{equation} \label{eq:streamdiff}
\tilde{\mathbf{x}}_{k}
\approx \left(\prod_{l=1}^{k-1}{(\lambda'_{k-l}\mathbf{H}^{\mathrm{res}}_{k-l} +  \tilde{\mathbf{A}}_{k-l})}\right)\tilde{\mathbf{x}}_{1}.
\end{equation}
As $\tilde{\mathbf{A}}_l$ is a finite-dimensional linear operator, there exists a Lipschitz constant $\ell_l = \sup_{\mathbf{Z} \ne \mathbf{0}} \frac{\|\tilde{\mathbf{A}}_l \mathbf{Z}\|_F}{\| \mathbf{Z} \|_F}$ for every layer $l$ which satisfies $\|\tilde{\mathbf{A}}_l \mathbf{Z}\|_F\le \ell_l\|\mathbf{Z}\|_F$ for $\forall \mathbf{Z} \in \mathbb{R}^{n \times d}$. 
Since $\|\mathbf{H}^{\mathrm{res}}_l\|_2\le 1$, the approximate recurrence in \eqref{eq:streamdiffiter} yields the following bound:
\begin{align} \label{eq:streamdiffnorm}
\Vert\tilde{\mathbf{x}}_{l+1}\Vert_F &\le \Vert\lambda'_l\mathbf{H}^{\mathrm{res}}_l\tilde{\mathbf{x}}_l\Vert_F+\Vert\tilde{\mathbf{A}}_l\tilde{\mathbf{x}}_l\Vert_F \le (\lambda'_l+\ell_l)\|\tilde{\mathbf{x}}_l\Vert_F \\
\quad \Rightarrow \quad 
\Vert\tilde{\mathbf{x}}_k\Vert_F &\le \left(\prod_{l=1}^{k-1}(\lambda'_l+\ell_l)\right)\Vert\tilde{\mathbf{x}}_1\Vert_F.
\end{align}
\end{proof}

\subsection{Proof of Theorem \ref{thm:alphagrad}}  \label{app:alphagradproof}

\noindent\textbf{Theorem \ref{thm:alphagrad}} (Gradient difference of different permutations). 
{\it 
For two different permutation matrices $\mathbf P_i,\mathbf P_j \in \mathbb{R}^{n \times n}$, the gradient difference of the pretraining loss $\mathcal L$ w.r.t. their coefficients $\alpha_{l,i},\alpha_{l,j}$ is bounded by
\begin{equation}
\vert (\nabla_{\alpha_{l,i}}-\nabla_{\alpha_{l,j}})\mathcal L \vert \le 2\lambda'_l \Vert \nabla_{\tilde{\mathbf x}_{l+1}}\mathcal L \Vert_F \Vert \tilde{\mathbf x}_l \Vert_F.
\end{equation}
}

\begin{proof}
From \eqref{eq:streamdiffiter}, the gradient of the stream difference at layer $l$ is
\begin{equation} \label{eq:streamdiffgrad}
\nabla_{\tilde{\mathbf x}_l}\mathcal L
=
(\lambda'_l\mathbf{H}^{\mathrm{res}}_l+\tilde{\mathbf{A}}_l)^\top
\nabla_{\tilde{\mathbf x}_{l+1}}\mathcal L \quad \Rightarrow \quad \nabla_{\tilde{\mathbf x}_{l+1}}\mathcal L
=
\left(\prod_{k=l+1}^{L-1}(\lambda'_k\mathbf{H}^{\mathrm{res}}_k+\tilde{\mathbf{A}}_k)\right)^\top
\nabla_{\tilde{\mathbf x}_{L}}\mathcal L.
\end{equation}
This shows that all later contractions $\lambda'_k$ with $k>l$ affect the gradient received by layer $l$.

We then analyze the update of $\boldsymbol{\alpha}$. From \eqref{eq:mhclite} and \eqref{eq:streamdiffiter}, we have
\begin{equation}
\frac{\partial\tilde{\mathbf x}_{l+1}}
{\partial\alpha_{l,m}}
=
\lambda'_l
\frac{\partial\mathbf{H}^{\mathrm{res}}_l}
{\partial\alpha_{l,m}}
\tilde{\mathbf x}_l
+
\underbrace{
\frac{\partial\tilde{\mathbf A}_l}
{\partial\alpha_{l,m}}
\tilde{\mathbf x}_l
}_{0}
= \lambda'_l \mathbf P_m \tilde{\mathbf x}_l.
\end{equation}
By the multivariate chain rule,
\begin{equation}
\frac{\partial\mathcal L}{\partial \alpha_{l,m}}
=
\sum_{i,j}
\frac{\partial\mathcal L}
{\partial(\tilde{\mathbf x}_{l+1})_{ij}}
\frac{\partial(\tilde{\mathbf x}_{l+1})_{ij}}
{\partial\alpha_{l,m}}
=
\lambda'_l
\sum_{i,j}
\left(\nabla_{\tilde{\mathbf x}_{l+1}}\mathcal L\right)_{ij}
\left(\mathbf P_m\tilde{\mathbf x}_l\right)_{ij},
\end{equation}
and therefore
\begin{equation} \label{eq:alphagraditer}
\nabla_{\alpha_{l,m}}\mathcal L
=
\lambda'_l \left\langle \nabla_{\tilde{\mathbf x}_{l+1}}\mathcal L,\ \mathbf P_m\tilde{\mathbf x}_l \right\rangle_F,
\end{equation}
where $\langle \mathbf{U}, \mathbf{V} \rangle_F := \sum_{i,j}{u_{ij}v_{ij}}$ is the Frobenius inner product.
Substituting \eqref{eq:streamdiffgrad} and \eqref{eq:streamdiff} into \eqref{eq:alphagraditer}, we obtain 
\begin{equation}
\nabla_{\alpha_{l,m}}\mathcal L
=
\lambda'_l \left\langle 
{
\overbrace{
\left(\prod_{k=l+1}^{L-1} (\lambda'_k\mathbf{H}^{\mathrm{res}}_k+\tilde{\mathbf{A}}_k)\right)^\top
\nabla_{\tilde{\mathbf x}_{L}}\mathcal L
}^{\lambda'_{>l}\text{ part: backward constraint}}
,\
\mathbf P_m 
\overbrace{
\left(\prod_{k=1}^{l-1}(\lambda'_{k-l}\mathbf{H}^{\mathrm{res}}_{k-l}+\tilde{\mathbf{A}}_{k-l})\right)
\tilde{\mathbf x}_1
}^{\lambda'_{<l}\text{ part: forward constraint}}
} \right\rangle_{F}.
\end{equation}
This indicates that the update of $\alpha_{l,m}$ is affected by three factors: (1) the contraction at the current layer $\lambda'_l$; (2) the contractions in deeper layers $\{\lambda'_{>l}\}$ introduced by backpropagation; and (3) the contractions in earlier layers $\{\lambda'_{<l}\}$ introduced by the forward process.

Now we compute the gradient difference between any $\alpha$-pair permutation matrices. For two different permutation matrices $\mathbf P_i$ and $\mathbf P_j$, we have
\begin{align}
(\nabla_{\alpha_{l,i}}-\nabla_{\alpha_{l,j}})\mathcal L &= \lambda'_l \left( \langle \nabla_{\tilde{\mathbf x}_{l+1}}\mathcal L,\ \mathbf P_i\tilde{\mathbf x}_l \rangle_F - \langle \nabla_{\tilde{\mathbf x}_{l+1}}\mathcal L,\ \mathbf P_j\tilde{\mathbf x}_l \rangle_F \right) \\
&=
\lambda'_l \left\langle \nabla_{\tilde{\mathbf x}_{l+1}}\mathcal L,\ (\mathbf P_{i} - \mathbf P_{j})\tilde{\mathbf x}_l \right\rangle_F.
\end{align}
By the Cauchy-Schwarz inequality,
\begin{equation}
\vert(\nabla_{\alpha_{l,i}}-\nabla_{\alpha_{l,j}})\mathcal L\vert \le \lambda'_l \Vert\nabla_{\tilde{\mathbf x}_{l+1}}\mathcal L \Vert_F \Vert(\mathbf P_{i}-\mathbf P_{j})\tilde{\mathbf x}_l \Vert_F.
\end{equation}
Since a permutation matrix is orthogonal, left multiplication by it preserves the Frobenius norm. With $\Vert \mathbf P_{i}-\mathbf P_{j} \Vert_2 \le \Vert \mathbf P_{i} \Vert_2 + \Vert \mathbf P_{j} \Vert_2 =2$, we reach the final result:
\begin{equation}
\vert (\nabla_{\alpha_{l,i}}-\nabla_{\alpha_{l,j}})\mathcal L \vert \le 2\lambda'_l \Vert \nabla_{\tilde{\mathbf x}_{l+1}}\mathcal L \Vert_F \Vert \tilde{\mathbf x}_l \Vert_F.
\end{equation}
\end{proof}

\section{Experiment Setup} \label{app:setup}

\paragraph{Datasets.}
OpenWebText~\citep{openwebtext} is built from Reddit-linked webpages, and FineWeb-Edu~\citep{finewebedu} is an education-oriented subset of FineWeb. For OpenWebText, we first tokenize it with GPT-2 BPE tokenizer and use a small validation split; this gives $\sim\!\!9.04$B training tokens and $\sim\!\!4.43$M validation tokens after preprocessing. 
For FineWeb-Edu, we use the ``sample-10BT'' split which contains roughly 10B GPT-2 tokens, and construct a validation split following the same pipeline. This gives $\sim\!\!9.90$B training tokens and $\sim\!\!49.10$M validation tokens.

\paragraph{Hardware and hyperparameters.}
The nanoGPT experiments are conducted on 8$\times$ 80GB NVIDIA H100 GPUs. For these comparative experiments, all models are pretrained for $10$k iterations. For a fair comparison, parameter initializations and nanoGPT-related hyperparameters of \model strictly follow the configurations in \texttt{mHC-lite}~\citep{mhclite}. 

\subsection{Experiment Setup for Large-Scale MoEs}
\label{app:large-scale-setup}

\paragraph{Architecture.}
Table~\ref{tab:large-scale-architecture} summarizes the 6B and 300B configurations. Here, depth counts Transformer blocks; each block contains separate attention and feed-forward hyper-connections. Both models use bias-free linear projections, SwiGLU feed-forward networks~\citep{swiglu}, RMSNorm~\citep{rmsnorm} with $\epsilon=10^{-6}$, query/key normalization, and zero attention and hidden dropout. Both use rotary position embeddings~\citep{rope} with base $10{,}000$; the 300B command explicitly selects YaRN~\citep{yarn} with both magnitude scaling factors set to $1$. The 6B configuration additionally enables one multi-token prediction (MTP) layer with MTP auxiliary loss weight $0.1$ and hyper-connections disabled.

\paragraph{Routing.}
Both configurations use sigmoid routing scores, expert bias updates, and sequence-level auxiliary load-balancing loss with coefficient $10^{-3}$. The expert bias update rates are $10^{-4}$ for 6B and $10^{-3}$ for 300B, and the scaling factors for routed outputs are $2.0$ and $2.5$, respectively.

\paragraph{Hyper-connection settings.}
Both runs enable \model with $n=4$ residual streams. In the 6B implementation, we use $\lambda=0.97$ and $K=4$. For 300B, the run uses $\lambda=0.96$ and $K=6$.  
For \texttt{mHC}, both configurations specify 20 Sinkhorn iterations.

\paragraph{Data and optimization.}
Both runs use tokenized FineWeb-Edu data. Table~\ref{tab:large-scale-training} reports training steps. Token counts are computed as completed steps multiplied by global batch size and sequence length; they count processed token positions. 

\begin{table}[H]
\centering
\small
\caption{Architecture settings for the large-scale experiments. Expert widths are intermediate dimensions. Model scales follow the experiment designations.}
\label{tab:large-scale-architecture}
\begin{tabular}{lrr}
\toprule
Setting & 6B-MoE & 300B-MoE \\
\midrule
Transformer blocks & 28 & 48 \\
Hidden size & 2,048 & 7,168 \\
Dense feed-forward width & 5,632 & 18,432 \\
Initial dense blocks & 2 & 3 \\
MoE layers & 26 & 45 \\
Attention heads & 16 & 128 \\
Query low-rank dimension & 512 & 1,536 \\
Key/value low-rank dimension & 256 & 512 \\
Non-rotary query/key head dimension & 128 & 128 \\
Rotary query/key head dimension & 64 & 64 \\
Value head dimension & 128 & 128 \\
Num of experts & 64 & 256 \\
Selected routed experts per token & 4 & 6 \\
Routed expert width & 512 & 1,152 \\
Shared expert intermediate width (total) & 1,024 & 1,152 \\
Vocabulary size & 131,072 & 131,072 \\
Context length & 4,096 & 4,096 \\
Residual streams & 4 & 4 \\
\bottomrule
\end{tabular}
\end{table}


\begin{table}[H]
\centering
\small
\caption{Training settings and completed training lengths. Token counts use the completed steps. A dash denotes a value not specified in the supplied configuration.}
\label{tab:large-scale-training}
\begin{tabular}{lrr}
\toprule
Setting & 6B-MoE & 300B-MoE \\
\midrule
Optimizer & AdamW & Muon \\
Learning rate after warmup & $10^{-5}$ & $2\times10^{-4}$ \\
Warmup steps & 2,000 & 200 \\
Schedule after warmup & Constant & Constant \\
Configured training steps & 71,500 & 2,000 \\
Global batch size & 512 & 4,096 \\
Micro-batch size & 1 & 1 \\
Tokens per optimization step & 2,097,152 & 16,777,216 \\
Processed tokens & $149.95$B & $33.55$B \\
Gradient clipping norm & 1.0 & 1.0 \\
Optimizer $\beta_1,\beta_2$ settings & $(0.9,0.95)$ & $(0.9,0.95)$ \\
AdamW $\epsilon$ & $10^{-8}$ & --- \\
Main computation precision & BF16 & BF16 \\
Tensor/context parallel degree & $1/1$ & $1/1$ \\
Pipeline parallel degree & 1 & 10 \\
Expert parallel degree & 8 & 8 \\
\bottomrule
\end{tabular}
\end{table}

\begin{table*}[!t]
\centering
\caption{nanoGPT-S/M results with seed 1338 (mean $\pm$ standard deviation for 3 replicate experiments). All methods use 4 streams. \textit{Avg. Loss Rank} marks the average rank over the Train Loss and Val. Loss. 
Best results are in \textbf{bold}. }
\label{tab:seed_results}
\begin{threeparttable}
\renewcommand{\arraystretch}{1.12}
\begin{adjustbox}{width=0.9\textwidth,center}
\begin{tabular}{lccccc}
\toprule[0.12em]
& \multicolumn{2}{c}{nanoGPT-S, \#Layers=12} & \multicolumn{2}{c}{nanoGPT-M, \#Layers=24} & \\
\cmidrule(lr){2-3}\cmidrule(lr){4-5}
Method & Train Loss & Val. Loss & Train Loss & Val. Loss & \textit{Avg. Loss Rank} \\
\midrule
\rowcolor{gray!0.9} \multicolumn{6}{c}{\textit{OpenWebText}} \\
\texttt{mHC} & $3.4677_{\pm 0.0013}$ & $3.4822_{\pm 0.0009}$ & $3.2674_{\pm 0.0021}$ & $3.2699_{\pm 0.0019}$ & $3.1$ \\
\texttt{mHC-lite} & $3.4668_{\pm 0.0002}$ & $3.4808_{\pm 0.0003}$ & $3.2731_{\pm 0.0022}$ & $3.2753_{\pm 0.0018}$ & $3.0$ \\
\texttt{KromHC} & $3.4672_{\pm 0.0027}$ & $3.4818_{\pm 0.0027}$ & $3.2677_{\pm 0.0016}$ & $3.2699_{\pm 0.0012}$ & $2.9$ \\
\rowcolor{violet!10}
\textbf{\model}, $\lambda=0.97$, $K=2$ & $\mathbf{3.4651}_{\pm 0.0016}$ & $\mathbf{3.4795}_{\pm 0.0014}$ & $\mathbf{3.2626}_{\pm 0.0034}$ & $\mathbf{3.2653}_{\pm 0.0041}$ & $\mathbf{1.0}$ \\

\midrule
\rowcolor{gray!0.9} \multicolumn{6}{c}{\textit{FineWeb-Edu}} \\
\texttt{mHC} & $\mathbf{3.4638}_{\pm 0.0016}$ & $\mathbf{3.4774}_{\pm 0.0017}$ & $3.2402_{\pm 0.0016}$ & $3.2531_{\pm 0.0018}$ & $2.1$ \\
\texttt{mHC-lite} & $3.4644_{\pm 0.0020}$ & $3.4779_{\pm 0.0021}$ & $3.2399_{\pm 0.0022}$ & $3.2526_{\pm 0.0020}$ & $3.0$ \\
\texttt{KromHC} & $\mathbf{3.4638}_{\pm 0.0002}$ & $3.4775_{\pm 0.0004}$ & $3.2430_{\pm 0.0032}$ & $3.2556_{\pm 0.0030}$ & $2.9$ \\
\rowcolor{violet!10}
\textbf{\model}, $\lambda=0.97$, $K=2$ & $3.4641_{\pm 0.0007}$ & $3.4777_{\pm 0.0008}$ & $\mathbf{3.2385}_{\pm 0.0003}$ & $\mathbf{3.2512}_{\pm 0.0007}$ & $\mathbf{2.0}$ \\
\bottomrule[0.12em]
\end{tabular}
\end{adjustbox}
\end{threeparttable}
\end{table*}

\vspace{1em}

\begin{figure*}[!t]
  \centering
  \includegraphics[width=\textwidth]{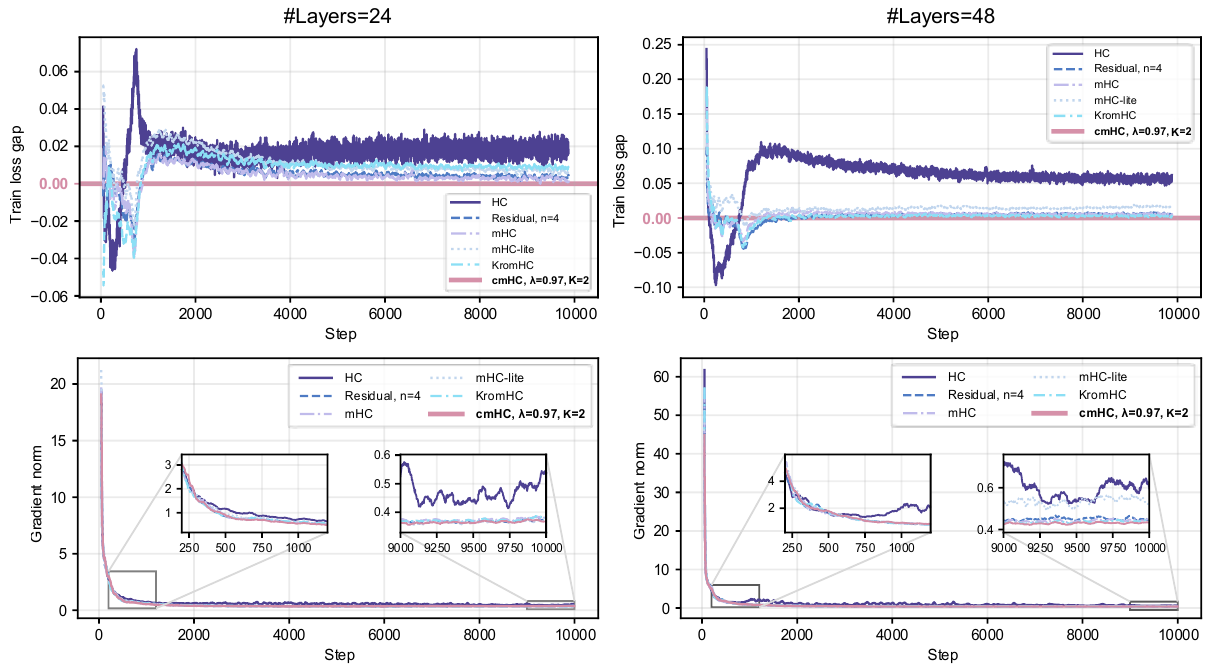}
  \caption{Loss gaps relative to \model and gradient norms (nanoGPT-M and nanoGPT-L; OpenWebText).}
  \label{fig:lossandnorm}
\end{figure*}

\section{Full Results} \label{app:experiments}

\subsection{Results on Seed 1338} \label{app:seed}

To further show that the observed improvements stem from gains in model capability rather than a lucky seed, we also repeat the nanoGPT-S/M experiments with seed 1338, in addition to default seed 1337 used for Table~\ref{tab:main_lm_results}. We show results of \model and other strong baselines in Table~\ref{tab:seed_results}. 

With seed 1338, \model has the lowest mean training and validation losses among the evaluated baselines on nanoGPT-S/M + OpenWebText and nanoGPT-M + FineWeb-Edu. Its train/validation loss gaps to the strongest evaluated baseline are 0.0017/0.0013, 0.0048/0.0046 and 0.0014/0.0014, respectively. Besides, 5/6 of the gaps exceed \model's corresponding standard deviations. These are also the same three settings in which \model has the lowest mean validation loss in Table~\ref{tab:main_lm_results}, providing evidence that the advantage of \model persists across seeds. On nanoGPT-S + FineWeb-Edu, the setting that \texttt{mHC-lite} outperforms \model by 0.0013/0.0012 in Table~\ref{tab:main_lm_results}, all four methods perform on par: their mean training and validation losses span only 0.0006 and 0.0005, respectively.

\begin{figure*}[!t]
  \centering
  \includegraphics[width=\textwidth]{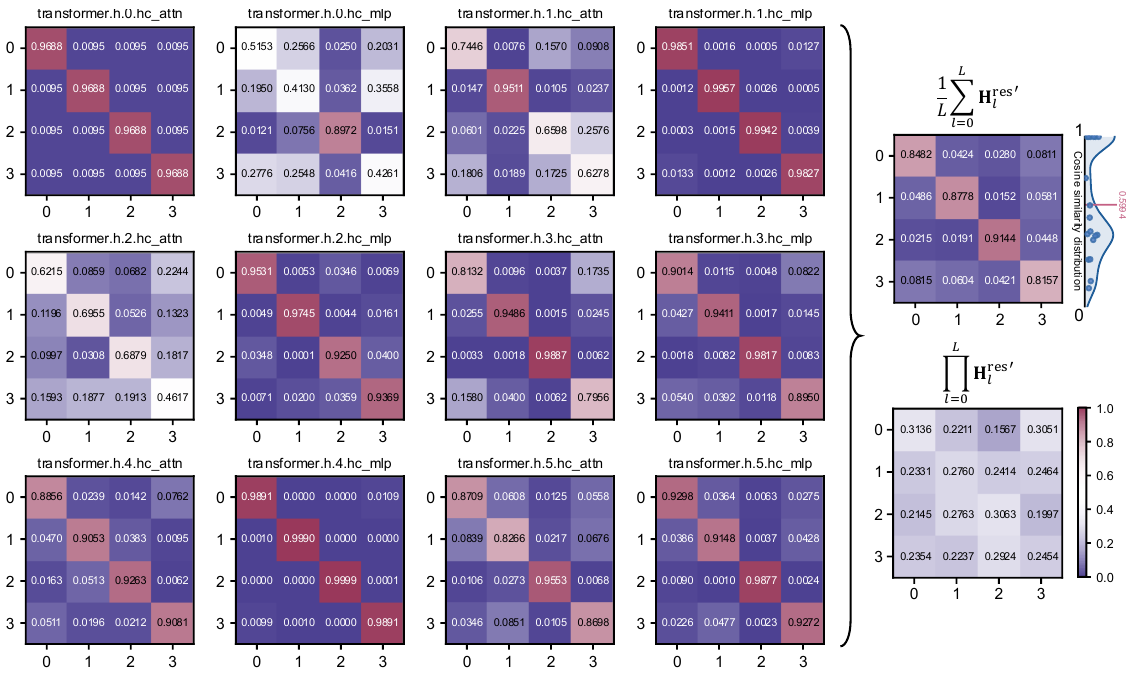}
  \caption{Full heatmaps of residual matrices (\model, $\lambda=0.97$, $K=2$; nanoGPT-L). \textit{Left}: residual matrices $\mathbf{H}^{\mathrm{res}}_l$ for every $l$. \textit{Right}: average and product of residual matrices, and cosine-similarity distribution between streams.}
  \label{fig:heatmapall}
\end{figure*}

\subsection{Training Loss and Gradient Norm on GPT-2} \label{app:gpt2}

Figure~\ref{fig:lossandnorm} shows full results for nanoGPT-M and nanoGPT-L, similar to Figure~\ref{fig:lossgrad}. The loss gap is measured relative to \model, so a positive value indicates a higher loss than ours. \model maintains a lower training loss than all compared residual and hyper-connection baselines at both depths. 
The gradient norms indicate that \model preserves the training stability of \texttt{mHC} in deeper models.

\subsection{Visualization of Residual Matrices} \label{app:hres}

Figure~\ref{fig:heatmapall} visualizes every residual matrix of the attention and MLP layers in \model. 
Despite that the contraction is only applied to the first two layers, we can observe the dynamic stream mixing in later layers like layer 3 and 5. Deeper layers also participate in stream mixing, but most of them remain close to near-identity mappings. This behavior matches our expectation: shallow-layer mixing has a stronger cumulative effect on the network, while allowing later layers to stay near-identity helps prevent over-mixing. The aggregated views further support this point: as the average residual matrix is near identity, the product of residual matrices is already close to, but still distinguishable from, the uniform barycenter $\mathbf{J}_n$. This contrast between the average and product residual matrices illustrates the mixing trade-off: encouraging stream mixing while keeping the streams distinguishable.

\subsection{Smoothness of the Mixing Coefficient} \label{app:alpha}

\begin{figure*}[!t]
  \centering
  \includegraphics[width=\textwidth]{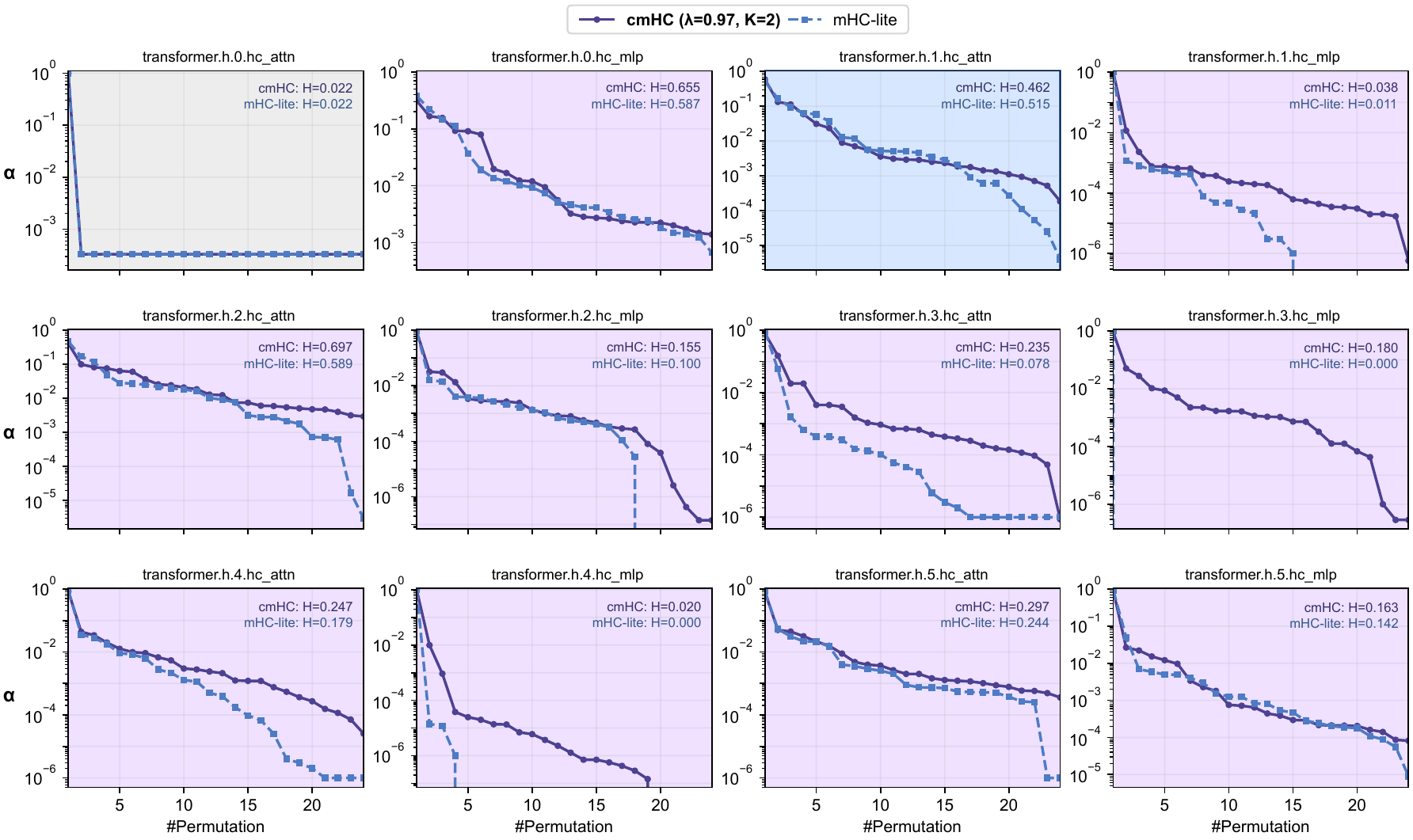}
  \caption{Full experiment results: distributions of mixing coefficients $\boldsymbol{\alpha}_l$ over the vertices of $\mathscr{B}_4$ for every $l$ (nanoGPT-L). A purple (blue) background marks cases where the entropy $H(\boldsymbol{\alpha}_l)$ of \model is larger (smaller) than that of $\texttt{mHC-lite}$.}
  \label{fig:alphadistall}
\end{figure*}

This section shows the original full version of Figure~\ref{fig:alphadistrank}. We further report the \textit{normalized entropy} $H(\boldsymbol{\alpha}_l) \in [0,1]$ to measure the smoothness of the distribution:
\begin{equation}
H(\boldsymbol{\alpha}_l) = \frac{-\sum_{m=1}^{n!}\alpha_{l,m}\log \alpha_{l,m}}{\log n!},
\end{equation}
where a larger value indicates a smoother distribution. 
Figure~\ref{fig:alphadistall} shows the distributions of $\boldsymbol{\alpha}_l$ and their normalized entropy for every layer $l$. In the majority of layers, \model assigns more mass to the non-dominant vertices and achieves higher normalized entropy than \texttt{mHC-lite}, despite that only the first 2 layers are contracted. 
This supports our conclusion that even a numerically small geometric contraction can induce a smoother $\boldsymbol{\alpha}$ distribution in deeper layers.

\vspace{1em}

\begin{figure*}[!t]
  \centering
  \includegraphics[width=\textwidth]{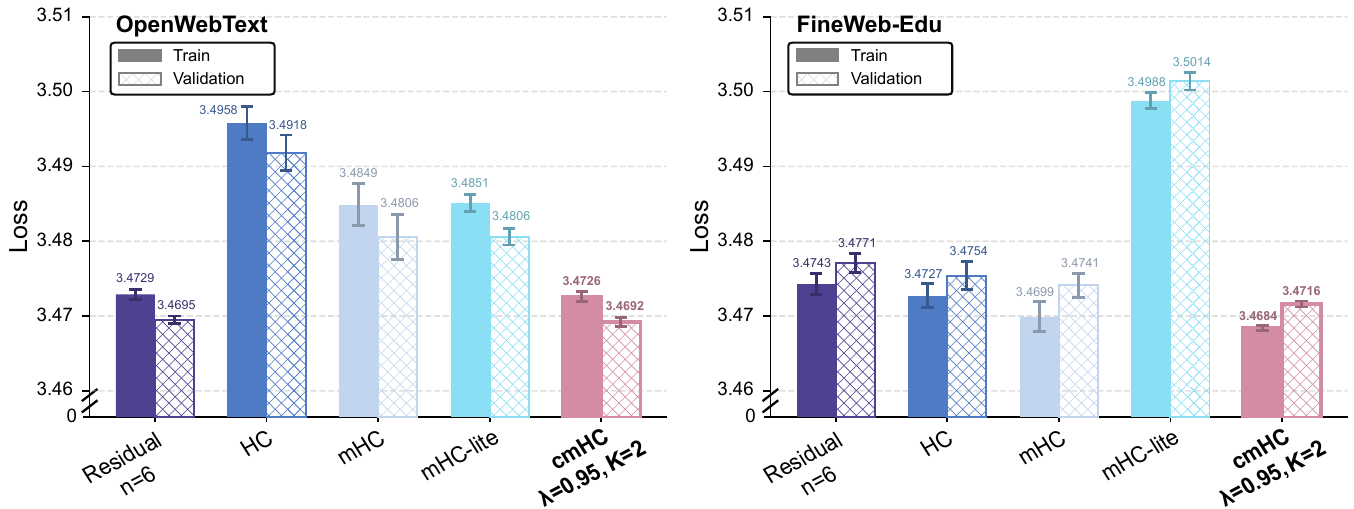}
  \caption{Comparative results with 6 residual streams (nanoGPT-S).}
  \label{fig:stream6}
\end{figure*}

\subsection{Scaling to More Streams}

Figure~\ref{fig:stream6} shows our experiments when scaling nanoGPT models to more hyper-connection streams. We use $\lambda=0.95$ such that the non-diagonal entries in $\mathbf{P}_m'$ remain $0.01$ like $\lambda=0.97$ for 4 streams.  
\model keeps the superior performance on both datasets when streams are expanded to 6.

\section{Implementation Details of the Fused Operator}
\label{app:fused-operator}

To reduce the execution overhead of \model, we develop an efficient fused-operator implementation. This
section describes its design and implementation and compares its operator-level performance with the \texttt{mHC}
fused-operator pipeline and an equivalent unfused primitive-operator implementation.

\begin{figure}[t]
  \centering
  \includegraphics[width=\linewidth,keepaspectratio]{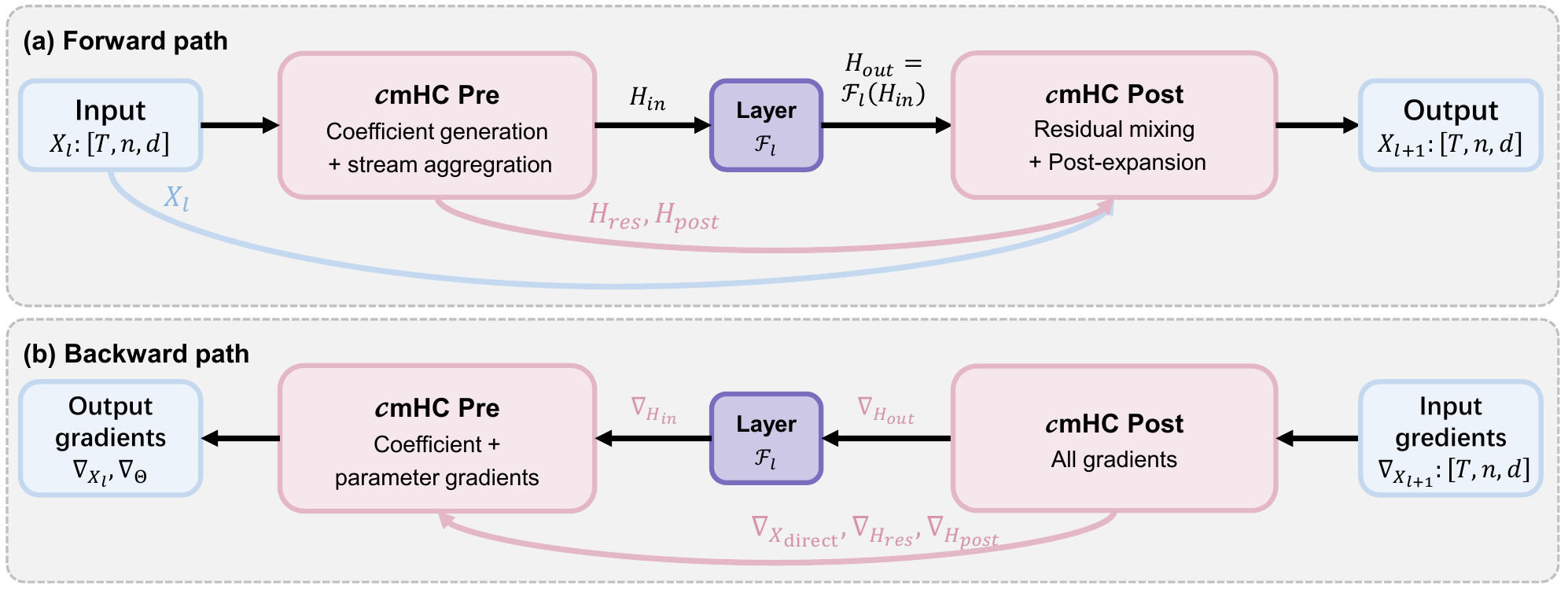}
  \caption{Fused-operator pipeline for \model. The backward path reverses the forward order and routes direct
  residual-path gradients from Post backward to Pre backward.}
  \label{fig:cmhc-pipeline}
\end{figure}

We implement the fused operators in Ascend C, a C/C++-compatible programming language and API system for
developing custom kernels on Ascend NPUs~\citep{huawei2025ascendc}. A \model layer places two fused operator
boundaries around the model sublayer. The fused Pre operator produces the sublayer input and residual-path coefficients;
the fused Post operator then performs residual mixing and post expansion. Training executes the corresponding operators in
reverse order. The implementation contract is
\begin{equation}
  (H_{\mathrm{in}},H_{\mathrm{res}},H_{\mathrm{post}})
    = \operatorname{Pre}(X_l),\quad
  h_l^{\mathrm{out}}=\mathcal{F}_l(H_{\mathrm{in}}),\quad
  X_{l+1}=\operatorname{Post}(X_l,H_{\mathrm{res}},h_l^{\mathrm{out}},H_{\mathrm{post}}).
  \label{eq:operator-contract}
\end{equation}
Here the capitalized names follow the operator-interface convention:
$X_l\equiv\mathbf{x}_l$, $H_{\mathrm{res}}\equiv\mathbf{H}^{\mathrm{res}'}_l$,
$H_{\mathrm{in}}\equiv\mathbf{H}^{\mathrm{pre}}_l\mathbf{x}_l$, and
$H_{\mathrm{post}}\equiv\mathbf{H}^{\mathrm{post}}_l$. Thus these tensors are the implementation-level
counterparts of the quantities defined in \eqref{eq:hc} and \eqref{eq:contract}.
The model sublayer $\mathcal{F}_l$ remains outside the fusion boundary because it may be an attention or
feed-forward block. The resulting Pre and Post boundaries fuse only the surrounding connection logic, avoiding
repeated reads of the widened residual state and launches of small framework operators.
Figure~\ref{fig:cmhc-pipeline} shows the forward and backward organization.

\subsection{Architecture-Aware Fusion Principles}
\label{sec:architecture-aware-design}

Let $B$, $S$, $n$, and $D$ denote batch size, sequence length, residual-stream count, and per-stream hidden
dimension, and let $T=BS$. The implementation specializes $n=4$, while $T$ and $D$ are large. Accordingly,
large projections use cube cores, whereas reductions, nonlinearities, coefficient transformations, and
small-$n$ stream mixing use vector cores. Table~\ref{tab:architecture-affinity} summarizes this mapping.

\begin{table}[t]
  \centering
  \small
  \caption{Design principles and their mapping to the target architecture.}
  \label{tab:architecture-affinity}
  \begin{tabular}{p{0.25\linewidth}p{0.34\linewidth}p{0.32\linewidth}}
    \toprule[0.12em]
    Design consideration & Kernel realization & Intended effect \\
    \midrule
    Heterogeneous arithmetic
      & Dense projections on cube cores; reductions, nonlinearities, and scale-accumulate paths on vector cores
      & Match arithmetic granularity to the core type \\\midrule
    Small stream dimension
      & Use vector scaling, accumulation, and reduction for $n=4$ Post mixing
      & Avoid GEMM setup, padding, and format conversion for tiny matrices \\\midrule
    Limited on-chip storage
      & Tile token and hidden dimensions; retain small coefficients and accumulators in Unified Buffer (UB)
      & Increase on-chip reuse within the UB capacity \\\midrule
    Global-memory traffic
      & Fuse common-input operations; select ND/NZ workspace layouts at vector--cube handoffs; selectively bypass L2 for one-pass tensors
      & Reduce intermediate traffic, layout conversion, and cache pollution \\\midrule
    Mixed-precision execution
      & Keep activations in BF16/FP16 and coefficients and accumulators in FP32
      & Reduce bandwidth while preserving coefficient accuracy \\\midrule
    Launch and scheduling overhead
      & Produce all outputs of each Pre or Post direction in one launch
      & Amortize launch cost and avoid repeated state scans \\
    \bottomrule[0.12em]
  \end{tabular}
\end{table}

Memory-layout and cache policies are selected by input shape. Small workloads retain a dense ND workspace,
whereas sufficiently large $T$ and $D$ use a blocked NZ workspace at vector--cube handoffs. Large one-pass
tensors selectively bypass L2, while weights, cached bases, and intermediates reused across stages remain
cacheable. These conditional paths preserve useful locality for small inputs and avoid conversion and cache
pollution at larger scales.

Pre combines vector and cube cores because it contains both high-dimensional projections and coefficient
transformations. Post uses vector cores only: its $4\times4$ coefficient matrix cannot amortize a cube-core
schedule, whereas its $D$-wide scale--accumulate loops map directly to vector instructions.

\subsection{Fused Pre Forward Kernel}
\label{sec:forward-kernel}

For the Pre implementation, we use $D_{\mathrm{flat}}=nD$, $R=n!$, and $Q=R+2n$ with $n=4$, $R=24$ and $Q=32$.
Table~\ref{tab:operator-interface} summarizes the tensors that determine its forward data path.

\begin{table}[htbp]
  \centering
  \small
  \caption{Primary tensor interface of the fused \model Pre forward operator.}
  \label{tab:operator-interface}
  \begin{tabular}{llll}
    \toprule[0.12em]
    Tensor & Role & Shape & Data type \\
    \midrule
    $X_l$ & Multi-stream input & $[B,S,n,D]$ & BF16/FP16 \\
    $\Phi$ & Projection weight & $[Q,nD]$ & FP32 \\
    $s$ & Group-wise scale & $[3]$ & FP32 \\
    $b$ & Projection bias & $[Q]$ & FP32 \\
    $\mathbf{P}'_l$ & Contracted permutation matrix & $[R,n,n]$ & FP32 \\
    $H_{\mathrm{in}}$ & Aggregated sublayer input & $[B,S,D]$ & BF16/FP16 \\
    $H_{\mathrm{post}}$ & Post-mapping coefficient & $[B,S,n]$ & FP32 \\
    $H_{\mathrm{res}}$ & Residual-mixing matrix & $[B,S,n,n]$ & FP32 \\
    \bottomrule[0.12em]
  \end{tabular}
\end{table}

Let $\mathbf{P}_m$ denote the $m$-th permutation matrix and
$\mathbf{J}_n=\mathbf{1}_n\mathbf{1}_n^\top/n$ the uniform doubly stochastic matrix. Before kernel launch,
the framework uses the normalized contraction coefficient
$\lambda'_l=(n\lambda_l-1)/(n-1)$ from Sec.~\ref{sec:contract} to construct
\begin{equation}
  \mathbf{P}'_{l,m}=\lambda'_l\mathbf{P}_m+(1-\lambda'_l)\mathbf{J}_n,
  \qquad m=1,\ldots,R,
  \label{eq:kernel-contracted-bases}
\end{equation}
which is algebraically identical to \eqref{eq:contract}. Under the depth-aware strategy, $\lambda_l=1$
(and hence $\lambda'_l=1$) beyond the first $K$ layers. The resulting basis tensor $\mathbf{P}'_l$ is cached
by the framework and passed to the forward and backward Pre kernels; the fused kernels therefore receive
the contracted bases rather than the scalar $\lambda_l$.

\begin{figure}[t]
  \centering
  \includegraphics[width=\linewidth,keepaspectratio]{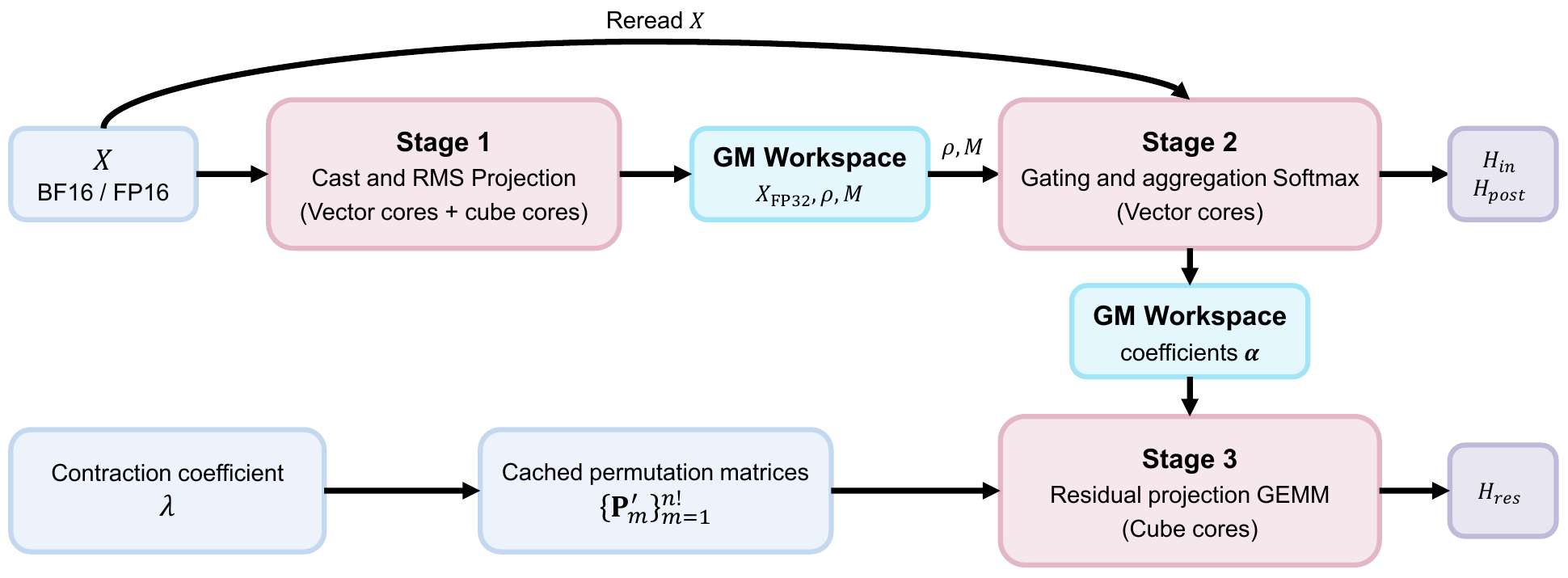}
  \caption{Three-stage data path of the fused \model Pre forward operator, including framework-side
  preparation of the contracted bases.}
  \label{fig:forward-pipeline}
\end{figure}

The operator exposes one entry and executes the three stages in Figure~\ref{fig:forward-pipeline}. Event
synchronization and GM workspace enforce producer--consumer ordering between vector and cube cores.

\paragraph{Stage 1: projection and RMS statistics.}
The vector cores cast BF16/FP16 input tiles to FP32 and write them to workspace. The same scan evaluates
\begin{equation}
  \rho=\left(\operatorname{mean}(X_l^2)+\epsilon_{\mathrm{norm}}\right)^{-1/2}
  \label{eq:rms-factor}
\end{equation}
for each token. The cube cores consume the converted workspace and compute
\begin{equation}
  M=\operatorname{FP32}(X_l)\Phi^{\mathsf T},
  \qquad M\in\mathbb{R}^{T\times Q}.
  \label{eq:forward-projection}
\end{equation}
Cross-core events prevent the projection from reading incomplete workspace tiles.

For sufficiently large workloads, the vector cores transform each FP32 tile from ND to NZ in UB before
writing the workspace. The cube cores then consume preformatted tiles through contiguous NZ transfers instead
of converting ND data on the read path. Host tiling retains the ND path for smaller inputs, where the
vector-side transformation cannot be amortized, and initializes only the selected matrix-multiplication
implementation. The NZ workspace aligns the $D_{\mathrm{flat}}$ dimension to the hardware block size and preserves a fixed
physical stride for tail tiles.

\paragraph{Stage 2: coefficient generation and stream aggregation.}
The vector cores apply group-wise scale and bias to $M$ and $\rho$, generating the Pre, Post, and
$R$-dimensional residual coefficients. A stable FP32 softmax processes the residual segment. The same stage
rereads low-precision $X_l$ and reduces its weighted streams into $H_{\mathrm{in}}$, avoiding separate gating,
softmax, and aggregation operators. For a token batch, the residual coefficients form
$\boldsymbol{A}\in\mathbb{R}^{T\times R}$, whose rows are the coefficient vectors
$\boldsymbol{\alpha}_l$ from \eqref{eq:mhclite}. They enter workspace for Stage 3 and are optionally exported
for backward reuse.

\paragraph{Stage 3: residual-matrix projection.}
After a barrier makes the coefficient tiles visible, the cube cores evaluate
\begin{equation}
  H_{\mathrm{res}}^{\flat}=\boldsymbol{A}(\mathbf{P}'_l)^{\flat},
  \qquad [T,R]\times[R,n^2]\rightarrow[T,n^2].
  \label{eq:residual-projection}
\end{equation}
For $n=4$, this is an FP32 $[T,24]\times[24,16]$ projection using the framework-cached bases. Work is
partitioned along the token dimension, with explicit tail handling.

The Pre path uses low-precision activation I/O and an FP32 coefficient path. $X_l$ and $H_{\mathrm{in}}$ remain
in BF16/FP16; projection weights, logits, softmax coefficients, and matrix multiplications use FP32. The cast
input, projection result, inverse RMS factor, and residual coefficients cross GM workspace only at required
vector-cube boundaries.

\subsection{Fused Pre Backward Kernel and State Reuse}
\label{sec:backward-kernel}

Pre backward fuses the gradients of stream aggregation, coefficient generation, softmax, and residual
projection. The vector cores reduce the upstream $H_{\mathrm{in}}$ gradient against the original streams and
prepare tiles for the projection-weight gradient. With $\nabla_H = \nabla_H \mathcal{L} \in\mathbb{R}^{T\times n^2}$ denoting the
residual-matrix gradient, the kernel evaluates
\begin{align}
  \nabla_{\boldsymbol{A}} &= \nabla_H((\mathbf{P}'_l)^{\flat})^{\mathsf T},
  \label{eq:grad-coeff}\\
  \nabla_Z &= \boldsymbol{A}\odot\left(\nabla_{\boldsymbol{A}}-\operatorname{reduce\_sum}(\boldsymbol{A}\odot \nabla_{\boldsymbol{A}})\right),
  \label{eq:grad-softmax}
\end{align}
where the reduction is independent for each token. The vector path reconstructs sigmoid derivatives from
saved Pre and Post outputs, propagates the inverse-RMS gradient, and writes $\nabla_M$ to workspace. The same
cached $\mathbf{P}'_l$ used in forward is read as a constant input; neither $\mathbf{P}'_l$ nor $\lambda_l$
receives a gradient in this operator.

The cube cores then execute the two high-dimensional matrix multiplications
\begin{equation}
  \nabla_X^{\mathrm{proj}}=\nabla_M\Phi,
  \qquad
  \nabla_{\Phi}=\nabla_M^{\mathsf T}\operatorname{FP32}(X_l).
  \label{eq:backward-gemms}
\end{equation}
The vector cores combine $\nabla_X^{\mathrm{proj}}$ with the aggregation, RMS, and optional direct Post gradients.
Gradients of $\Phi$, $s$, and $b$ are reduced deterministically; scale and bias use compensated
accumulation.

Training reuses the states in Table~\ref{tab:saved-states} instead of recomputing Pre. At $n=4$, they contain 65
FP32 values (260 bytes) per token. Because $H_{\mathrm{post}}$ is already a forward output, only the remaining
61 values are optional training outputs; inference disables them.

\begin{table}[t]
  \centering
  \small
  \caption{Forward states reused by the fused \model Pre backward operator for $n=4$.}
  \label{tab:saved-states}
  \begin{tabular}{lccr}
    \toprule[0.12em]
    State & Shape & Data type & FP32 values/token \\
    \midrule
    Inverse RMS factor & $[T,1]$ & FP32 & 1 \\
    Projection output $M$ & $[T,32]$ & FP32 & 32 \\
    Pre coefficients & $[T,4]$ & FP32 & 4 \\
    Post coefficients & $[T,4]$ & FP32 & 4 \\
    Residual coefficients $\boldsymbol{A}$ & $[T,24]$ & FP32 & 24 \\
    \midrule
    Total & -- & FP32 & 65 \\
    \bottomrule[0.12em]
  \end{tabular}
\end{table}

\subsection{Fused Post Forward and Backward Kernels}
\label{sec:post-kernels}

For token $t$, output stream $i$, and input stream $j$, Post computes
\begin{equation}
  X_{l+1,t,i,:}
  =H_{\mathrm{post},t,i}\,h^{\mathrm{out}}_{l,t,:}
   +\sum_{j=1}^{n}H_{\mathrm{res},t,j,i}\,X_{l,t,j,:}.
  \label{eq:post-forward}
\end{equation}
Input and output activations use BF16/FP16; $H_{\mathrm{res}}$ and $H_{\mathrm{post}}$ use FP32. The $n=4$
implementation supports $[B,S,n,D]$ and flattened $[T,n,D]$ layouts and requires $D$ to be divisible by 128.

Post forward uses vector cores only. Each core processes a token-hidden tile, initializes an FP32 accumulator
with $H_{\mathrm{post},i}h_l^{\mathrm{out}}$, and applies four residual-stream scale-accumulates. The result is
cast once on GM writeback. Small coefficients remain on chip, $h_l^{\mathrm{out}}$ is reused across output
streams, and double-buffered queues overlap transfer with computation.

Tiling selects between two reuse schedules. If all $n$ input tiles fit in UB, they remain resident across
output streams; otherwise, inputs are streamed while the FP32 accumulator remains resident. This avoids UB
overcommitment without changing the arithmetic.

Given the upstream gradient $\nabla_Y=\nabla_{X_{l+1}}$, the Post backward operator jointly evaluates
\begin{align}
  \nabla_{X_l,t,j,:} &= \sum_{i=1}^{n}H_{\mathrm{res},t,j,i}\nabla_{Y,t,i,:},
  &
  \nabla_{H_{\mathrm{res}},t,j,i} &= \sum_{d=1}^{D}X_{l,t,j,d}\nabla_{Y,t,i,d},
  \label{eq:post-grad-res}\\
  \nabla_{h^{\mathrm{out}},t,:} &= \sum_{i=1}^{n}H_{\mathrm{post},t,i}\nabla_{Y,t,i,:},
  &
  \nabla_{H_{\mathrm{post}},t,i} &= \sum_{d=1}^{D}h^{\mathrm{out}}_{l,t,d}\nabla_{Y,t,i,d}.
  \label{eq:post-grad-post}
\end{align}
All four gradient families are produced in one launch. Tokens are distributed across vector cores, and the
hidden dimension uses aligned tiles with an explicit tail path. The Post vector and residual matrix remain in
UB; vector multiply, multiply--accumulate, and reduction implement the small products. The coefficient
gradients accumulate in FP32 and are written once per token, while activation gradients are cast and written
per tile. Fusion reuses $\nabla_Y$, $X_l$, and the coefficients, avoiding four scans of $D$.

\subsection{Operator-Level Performance}
\label{sec:operator-performance}

We benchmark on an Ascend 910C server with CANN 9.1.0~\citep{huawei2026cann}. CANN (Compute Architecture for
Neural Networks) provides the operator library, compiler toolchain, and runtime for Ascend processors. The
\texttt{mHC} comparison operators are the fused implementations from this release. Both pipelines use the same Post
implementation, isolating their implementation difference to Pre.

All workloads use $B=1$, $S=4,096$, $n=4$, and $D\in\{2048,3584,7168\}$. Latency is reported in microseconds
by input shape. Speedup is $T_{\mathrm{baseline}}/T_{\mathrm{ours}}$. The model sublayer is excluded, and Pre
and Post were timed independently. The complete-pipeline totals and reductions are reported in
Table~\ref{tab:operator-main}; below we provide the detailed Pre/Post breakdowns.

Table~\ref{tab:pre-performance} isolates the differing Pre stage. The forward operator achieves $3.82\times$,
$2.51\times$, and $1.93\times$ speedups. The shape-aware layout and cache policies are most consequential at
large $D$, where wider rows otherwise increase cube-side layout-conversion cost and cache pressure. The
backward gain narrows with $D$, reaching $1.46\times$ at $D=7,168$ as projection gradients dominate
fixed-size coefficient work.

\begin{table}[t]
  \centering
  \small
  \caption{Latency of the fused \model Pre and \texttt{mHC} Pre operators.}
  \label{tab:pre-performance}
  \begin{tabular}{lrrcrrc}
    \toprule[0.12em]
    \multirow{2}{*}{\makecell{Input\\$[B,S,n,D]$}}
      & \multicolumn{2}{c}{Fwd latency ($\mu$s)}
      & \multirow{2}{*}{Fwd speedup}
      & \multicolumn{2}{c}{Bwd latency ($\mu$s)}
      & \multirow{2}{*}{Bwd speedup} 
      \\\cmidrule(){2-3} \cmidrule(){5-6}
      & \model & \texttt{mHC} & & \model & \texttt{mHC} & \\
    \midrule
    $[1,4096,4,2048]$ & $\mathbf{266.70}$ & $1,017.50$ & $3.82\times$ & $\mathbf{444.40}$ & $889.03$ & $2.00\times$ \\
    $[1,4096,4,3584]$ & $\mathbf{485.96}$ & $1,217.54$ & $2.51\times$ & $\mathbf{879.57}$ & $1,365.56$ & $1.55\times$ \\
    $[1,4096,4,7168]$ & $\mathbf{1,168.57}$ & $2,255.84$ & $1.93\times$ & $\mathbf{2,082.52}$ & $3,045.70$ & $1.46\times$ \\
    \bottomrule[0.12em]
  \end{tabular}
\end{table}

Table~\ref{tab:post-performance} reports the identical Post cost in both pipelines. It accounts for 51.8\%,
49.5\%, and 49.8\% of our forward-plus-backward cost, diluting the larger Pre-only speedups.

\begin{table}[t]
  \centering
  \small
  \caption{Latency of the fused Post forward and backward operators used in the complete pipeline.}
  \label{tab:post-performance}
  \begin{tabular}{lrrr}
    \toprule[0.12em]
    \makecell{Input\\$[B,S,n,D]$}
      & \makecell{Fwd latency\\($\mu$s)}
      & \makecell{Bwd latency\\($\mu$s)}
      & \makecell{Total latency\\($\mu$s)} \\
    \midrule
    $[1,4096,4,2048]$ & 109.32 & 654.17 & 763.49 \\
    $[1,4096,4,3584]$ & 160.20 & 1,180.29 & 1,340.49 \\
    $[1,4096,4,7168]$ & 993.84 & 2,227.63 & 3,221.47 \\
    \bottomrule[0.12em]
  \end{tabular}
\end{table}

\begin{table}[t!]
  \centering
  \small
  \caption{Latency of our fused Pre operator and the MindSpore primitive-operator composition. %
  }
  \label{tab:mindspore-performance}
  \begin{tabular}{lrrcrrc}
    \toprule[0.12em]
    \multirow{2}{*}{\makecell{Input\\$[B,S,n,D]$}}
      & \multicolumn{2}{c}{Fwd latency ($\mu$s)}
      & \multirow{2}{*}{Fwd speedup}
      & \multicolumn{2}{c}{Bwd latency ($\mu$s)}
      & \multirow{2}{*}{Bwd speedup} 
      \\\cmidrule(){2-3} \cmidrule(){5-6}
      & Ours & MindSpore & & Ours & MindSpore & \\
    \midrule
    $[1,4096,4,2048]$ & $\mathbf{314.17}$ & $1,524.50$ & $4.85\times$ & $\mathbf{620.95}$ & $3,042.44$ & $4.90\times$ \\
    $[1,4096,4,3584]$ & $\mathbf{567.43}$ & $2,607.34$ & $4.59\times$ & $\mathbf{1,259.12}$ & $5,455.29$ & $4.33\times$ \\
    $[1,4096,4,7168]$ & $\mathbf{1,252.86}$ & $5,325.69$ & $4.25\times$ & $\mathbf{3,517.94}$ & $11,119.79$ & $3.16\times$ \\
    \bottomrule[0.12em]
  \end{tabular}
\end{table}
Finally, Table~\ref{tab:mindspore-performance} compares our fused Pre operator with an unfused MindSpore
composition~\citep{mindspore2022whitepaper}. This isolates the benefit of custom-kernel fusion, while the
\texttt{mHC} comparison evaluates the contraction-specific design. The two comparisons are separate benchmark series;
absolute latencies for our operator are comparable only within each table.

Our fused Pre operator achieves $4.25\times$--$4.85\times$ forward and $3.16\times$--$4.90\times$ backward speedup over
the MindSpore composition, confirming that fusion removes substantial launch and intermediate memory
overhead. As $D$ grows, high-dimensional work and the identical Post cost reduce the complete-pipeline gain.

\stopcontents[appendix]

\end{document}